\documentclass{article}

\usepackage{PRIMEarxiv}

\usepackage[utf8]{inputenc} 
\usepackage[T1]{fontenc}    
\usepackage{hyperref}       
\usepackage{url}            
\usepackage{booktabs}       
\usepackage{amsfonts}       
\usepackage{nicefrac}       
\usepackage{microtype}      
\usepackage{lipsum}
\usepackage{fancyhdr}       
\usepackage{graphicx}       
\graphicspath{{media/}}     

\usepackage{microtype}
\usepackage{graphicx}
\usepackage{subfigure}
\usepackage{booktabs} 
\usepackage{amsmath}
\usepackage{float}
\usepackage{placeins}
\usepackage{amssymb}
\usepackage{verbatim}

\newcommand{\R}{\mathbb{R}}

\usepackage{bm}
\usepackage[dvipsnames]{xcolor}

\title{Do Kernel and Neural Embeddings help in Training and Generalization? \thanks{This work was partially supported by the Wallenberg AI, Autonomous Systems and Software Program (WASP) funded by the Knut and Alice Wallenberg Foundation as well as the Chalmers AI Research Centre. We would like to thank Ashkan Panahi and the anonymous reviewers for their constructive comments.
}
}

\author{
  Arman Rahbar, Emilio Jorge, Devdatt Dubhashi, Morteza Haghir Chehreghani \\
  Chalmers University of Technology  \\
  SE-412 96 G{\"o}teborg\\
  Sweden\\
  \texttt{\{armanr, emilio.jorge, dubhashi, morteza.chehreghani\}@chalmers.se} \\
}

\begin{document}
\maketitle







\begin{abstract}
Recent results on optimization and generalization properties of neural networks showed that in a simple two-layer network, the alignment of the labels to the eigenvectors of the corresponding Gram matrix determines the convergence of the optimization during training. Such analyses also provide upper bounds on the generalization error. We experimentally investigate the implications of these results to deeper networks via embeddings. We regard the layers preceding the final hidden layer as producing different representations of the input data which are then fed to the two-layer model. We show that these representations improve both optimization and generalization. In particular, we investigate three kernel representations when fed to the final hidden layer: the Gaussian kernel and its approximation by random Fourier features, kernels designed to imitate representations produced by neural networks and finally an optimal kernel designed to align the data with target labels. The approximated representations induced by these kernels are fed to the neural network and the optimization and generalization properties of the final model are evaluated and compared. 
\end{abstract}
\section{Introduction}
Deep neural network models \cite{BengioCV13,LeCunBH15,Schmidhuber15} provide the state-of-art results in several tasks and applications, although the theory has not been completely understood yet.
The well-known work of \cite{Zhang17} highlighted intriguing experimental phenomena about deep network training -- specifically, optimization and generalization -- and called for a rethinking of generalization in statistical learning theory. In particular, two fundamental questions that need to be answered are:
\\

{\bf Optimization.} Why do true labels give faster convergence rate than random labels for gradient descent?
\\

{\bf Generalization.} What property of properly labeled data controls generalization?
\\

An approach to addressing these questions was recently made by \cite{Arora19} in the context of a simple two--layer model by conducting a spectral analysis of the associated Gram matrix. They show that {\em training improves if the label vector aligns with the top eigenvectors} of the associated Gram matrix ($\mathbf H^{\infty}$ in section \ref{sec:spetheory}). In addition, they provide a data-dependent complexity measure which could be used to upper bound the generalization error of the neural network. This measure is also related to the Gram matrix.

However, their analysis applies only to a simple two layer network. \emph{How could their insights be extended to deeper networks? }

A widely held intuitive view is that deep layers generate expressive \emph{representations} of the raw input data. Adopting this view, one may consider a model where a representation generated by successive neural network layers is viewed as a \emph{kernel embedding} which is then fed into the two--layer model of \cite{Arora19}. The connection between neural networks and kernel machines has long been studied; \cite{CS09} introduced kernels that mimic deep networks and \cite{TKG18} showed kernels equivalent to certain feed--forward neural networks. Recently, \cite{Belkin18} also make the case that progress on understanding deep learning is unlikely to move forward until similar phenomena in classical kernel machines are recognized and understood. 

The fundamental question we address is: \emph{Do representations provided by  kernel embeddings help in training and generalization}? We address this question in the context of the simple model of \cite{Arora19} and their approach based on a spectral analysis of the associated Gram matrix. Specifically, we take a view of a multi--layer network as first producing an embedding $\phi(x)$ of the input which is then fed as input to the simple two layer network of \cite{Arora19}. While a general transformation $g(x)$ of the input data may have arbitrary effects, one might expect a universal kernel representation to improve performance. Do kernel representations create a better alignment of labels with the top eigenvectors of the Gram matrix? We investigate this first by using kernels which are label-unaware\footnote{We say Gaussian kernel is label-unaware because we do not employ any data labels to compute it, i.e., this kernel is not sensitive to the labels.} such as Gaussian kernel, using random Fourier features(RFF) to approximate the Gaussian kernel embedding \cite{RR07}.

Next, we address the question: \emph{Do representations provided by neural embeddings help in training and generalization?} We address this in two ways: first, we use kernels that have been designed to specifically mimic neural networks \cite{CS09}. Then we use data driven embeddings explicitly produced by the hidden layers in neural networks:  either using a subset of the same training data to compute such an embedding, or transfer the inferred embedding from a different (but similar) domain. 


We indeed find substantial improvements in both training and generalization using any of the above kernel representations.

Finally we ask the question: \emph{What is a good representation for optimization?} The work of \cite{Arora19} suggests seeking a representation which makes the label vector align best to the top eigenvectors of the associated Gram matrix. This connects to the problem of \emph{kernel--target alignment} \cite{Christ01,CortesMohri2012}. We use the criterion suggested in these papers as a proxy for aligning the label vector to the top eigenvectors and compute the representation so produced. Indeed we observe that this representation provides the best results for training. 

Thus this work shows that kernel and neural embeddings improve the alignment of target labels to the eigenvectors of the Gram matrix and thus help training. This suggests a way to extend the insights of \cite{Arora19} to deeper networks, and possible theoretical results in this direction. 

In addition, the work in \cite{Arora19} yields a data-dependent complexity measure to be used to upper bound the generalization error of a learned neural network. We adapt this measure to be used with kernel and neural embeddings to analyze the generalization performance resulted from these representations. 

\section{Spectral Theory}
\label{sec:spetheory}
\paragraph{Network model.}
In \cite{Arora19}, the authors consider a simple two layer network model:
\begin{equation}
    f_{\bm W,\bm a}(\bm x) = \frac{1}{\sqrt{m}}\sum_{r=1}^m a_r \max(0, \bm w_r^T \bm x_i), 
\end{equation}
with $\bm x\in \R^d$, $\bm w_1$, .. $\bm w_m \in \R^{d\times m}$ and ($a_1$, .. $ a_m)^T \in \R^m$ (where $m$ specifies the number of neurons in the hidden layer, i.e., its width). These can be written jointly as $\mathbf a=( a_1, ..,  a_m)^T$ and $\mathbf W=(\bm w_1, .. , \bm w_m)$. This network is trained on dataset of data points $\{x_i\}$ and their targets $\{y_i\}$.

They provide a fine grained analysis of training and generalization error by a spectral analysis of the \emph{Gram matrix}:
\begin{equation}
\label{gramorg}
\begin{aligned}
 \mathbf H^{\infty}_{i,j} := E_{\mathbf W\sim \mathcal{N}(0,\bm{\mathcal{I}})}[\mathbf x_i^T  \mathbf x_j 1\{ \mathbf w^T \mathbf x_i \geq 0, \mathbf w^T \mathbf x_j \geq 0\}] \\ = \frac{\mathbf x_i^T  \mathbf x_j(\pi-\arccos{(\mathbf x_i^T  \mathbf x_j)})}{2\pi}.
 \end{aligned}
\end{equation}
If 
\begin{equation}
    \mathbf H^{\infty}  = \sum_i \lambda_i \mathbf v_i \mathbf v^T_i
\end{equation} 
is the orthonormal decomposition of $\mathbf H^{\infty}$, \cite{Arora19} shows that training improves if the label vector $y$ aligns with the eigenvectors corresponding to the top eigenvalues of $\mathbf H^{\infty}$. 

We extend the two-layer ReLU network in \cite{Arora19} via adding different types of embeddings $\phi$ at the input layer corresponding to a kernel $\mathcal{K}$.
The corresponding model is:
\begin{equation}
\label{eq:model}
    f_{\mathbf{W},\mathbf{a}}(\mathbf{x}) = \frac{1}{\sqrt{m}}\sum_{r=1}^m a_r \max(0, \mathbf{w}_r^T \bm{\phi}(\mathbf{x}_i)).
\end{equation}

For a representation $\left(\bm \phi(\bm x_i),i \in [n]\right)$ corresponding to a kernel $\mathcal{K}$, Since $\mathcal{K}(\bm x_i,\bm x_j) = \phi(\bm x_i)^T\phi(\bm x_j),i,j \in [n])$, similar to \ref{gramorg} define the Gram Matrix
\begin{equation}
\begin{aligned}
    \mathbf H(\mathcal{K})^{\infty}_{i,j} := E_{\mathbf W}[\mathcal{K}(\bm x_i,\bm x_j) 1\{ \mathbf w^T \bm \phi(\bm x_i) \geq 0, \mathbf w^T \bm \phi(\bm x_j) \geq 0\}] \\
    = \frac{\mathcal{K}(\bm x_i,\bm x_j)(\pi-\arccos{(\mathcal{K}(\bm x_i,\bm x_j)})}{2\pi}.
\end{aligned}
\end{equation} 
and let its eigenvalues be ordered  as 
\begin{equation}
    \lambda_0(\mathcal{K}) \geq \lambda_1(\mathcal{K}) \geq \cdots \geq \lambda_{n-1}(\mathcal{K})
\end{equation} and let $\mathbf v_0(\mathcal{K}),\cdots, \mathbf v_{n-1}(\mathcal{K})$ be the corresponding eigenvectors. 

A kernel $\mathcal{K}$ such that the corresponding eigenvectors align well with the labels would be expected to perform well for training optimization. This is related to kernel target alignment \cite{Christ01}.

\paragraph{Optimization.}
By adapting the convergence of the gradient descent for basic model in \cite{Arora19}, the convergence of our kernelized network is controlled by 

%
\begin{equation}
 \sqrt{\sum_i (1 - \eta \lambda_i(\mathcal{K}))^{2k} (\mathbf v(\mathcal{K})_i^T \mathbf y)^2}
\end{equation}

\paragraph{Generalization.}
We can also adapt the generalization performance of  the simple two-layer network in \cite{Arora19} to the kernel embedding setting as

\begin{equation}
  \sqrt{
  \frac{ 2\mathbf y^T(\mathbf H(\mathcal{K})^\infty)^{-1}\mathbf y}{n}}
 \label{eq:gen_kernel}
\end{equation}

For both optimization and generalization, we replace the data features with the feature vectors induced by the kernel embeddings. 

\section{Numerical Analysis}
\label{sec:numerical}

\begin{figure*}[ht!]
    \centering
    \subfigure[MNIST training loss as a function of epoch number]
    {
        \includegraphics[width=0.45\textwidth]{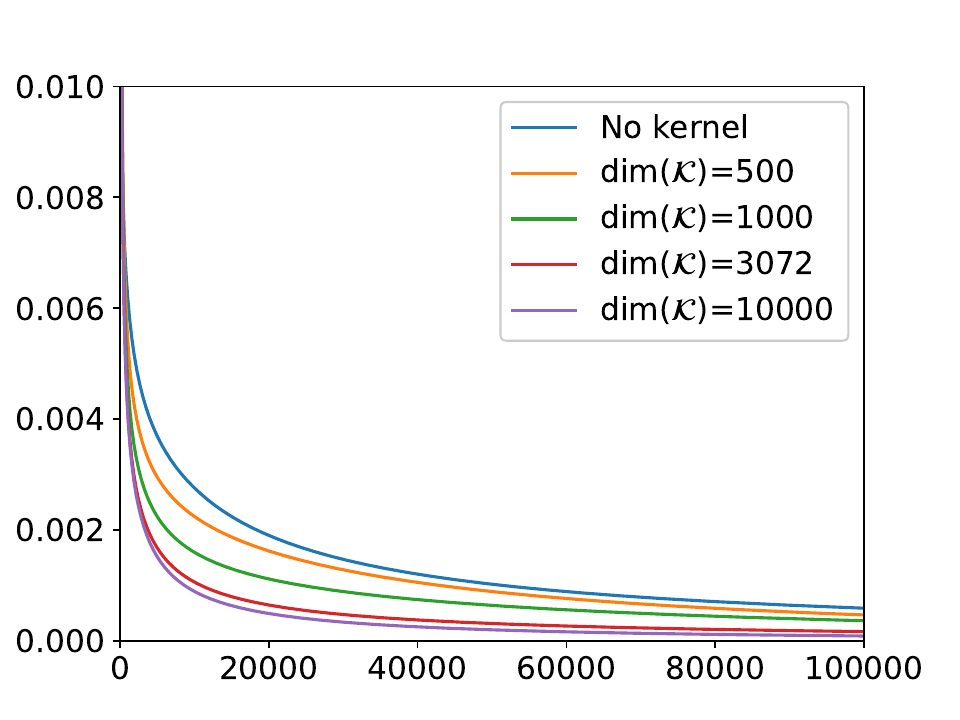}
        \label{fig:train_loss_mnist}
    }\hspace{3mm}
    \subfigure[CIFAR-10 training loss as a function of epoch number]
    {
        \includegraphics[width=0.45\textwidth]{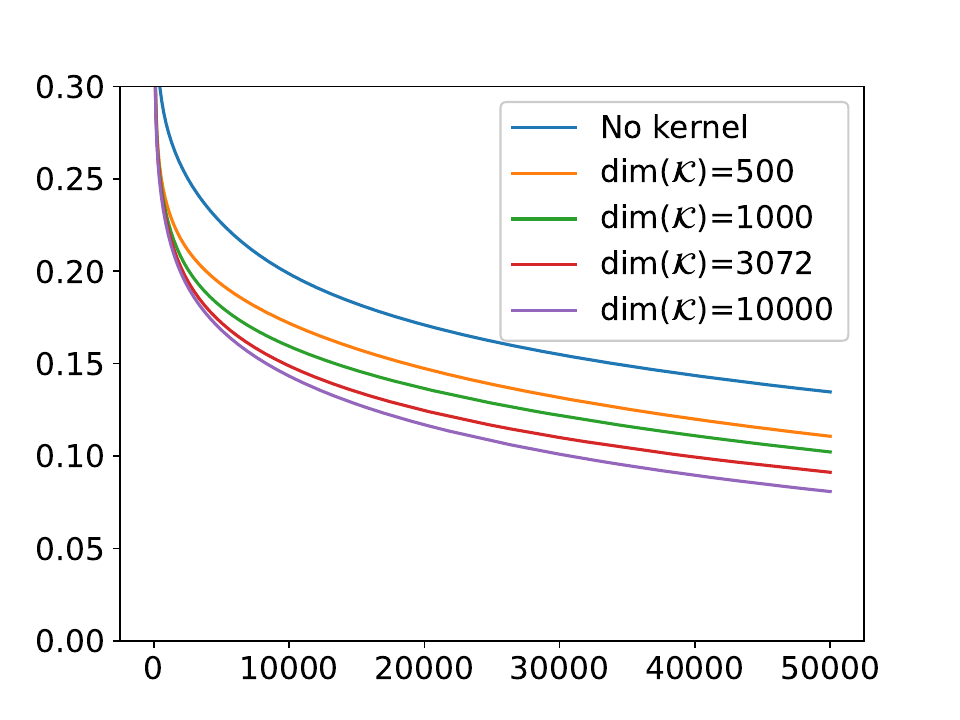}
        \label{fig:train_loss_cifar}
    }
    \subfigure[CIFAR-10 projections]
    {
        \includegraphics[width=0.45\textwidth]{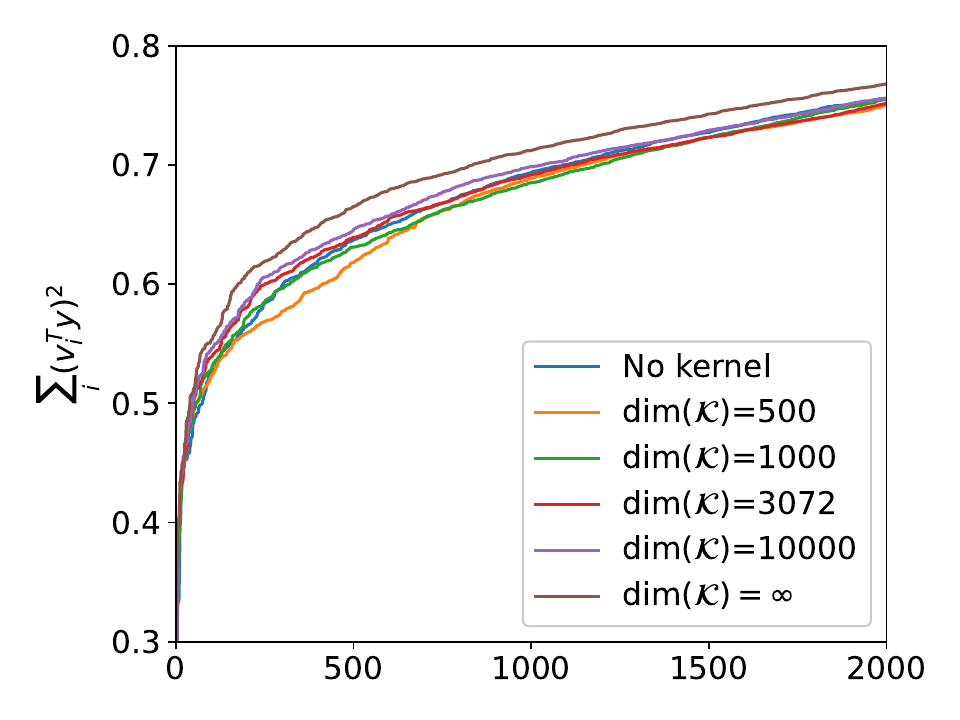}
        \label{fig:train_projections_loss_cifar}
    }
    \caption{Performance on first two classes of MNIST and CIFAR-10 as training datasets. We observe that the different kernels yield faster convergence of the loss function on training data compared to non-kernel variant. Figure \ref{fig:train_projections_loss_cifar} demonstrates alignment of top eigenvalues and the projections of true labels on corresponding eigenvectors.}
    \label{fig:train_loss}
\end{figure*}

We perform our numerical analysis on various datasets. In addition to MNIST and CIFAIR-10 which are used for the experiments in \cite{Arora19}, we used other datasets such as LSUN \cite{yu15lsun} and Fashion-MNIST in several parts of the paper. As in \cite{Arora19}, we only look at two classes when training our models, and set the label $y_i=+1$ if image $i$ belongs to the first class and $y_i=-1$ if it belongs to the second class. To follow the setting in \cite{Arora19}, the images are normalized such that $||\bm x_i||_2=1$. This is also done for the different kernel embeddings such that $||\bm \phi(\bm x_i)||_2=1$.
\label{data}

The weights in equation \eqref{eq:model} are initialized as follows:
\begin{equation}
    \bm w_i \sim \mathcal{N}(0, k^2 \bm{\mathcal{I}}), \, a_r \sim \text{Unif}(\{-1,1\}), \forall r \in [m].
\end{equation}

We then use the following loss function to  train the model to predict the image labels.
\begin{equation}
\Phi(\mathbf{W}, \mathbf{a}) = 1/2 \sum_{(i=1)}^n(y_i- f_{\mathbf{W},\mathbf{a}}(\mathbf{x}))^2
\end{equation}

For optimization, we use (full batch) gradient descent with the learning rate $\eta$. We set $k=10^{-2}, \eta=2 \times 10^{-4}$ similar to \cite{Arora19}. 

\subsection{Gaussian kernel method}
We first use the Gaussian kernel 
\begin{equation}
\label{eq:gaussian}
    \mathcal{K}(\bm x_i, \bm x_j) := \exp\left(- \gamma \|\bm x_i - \bm x_j\|^2\right).
\end{equation}
The corresponding embedding is infinite dimensional, hence we consider the fast approximations to the kernel given by \emph{random Fourier features} (RFF) \cite{RR07}. The idea of random Fourier features is to construct an explicit feature map which is of a dimension much lower than the number of observations, but  the resulting inner product approximates the desired kernel function. Specifically, instead of using the kernel trick for computing the inner product in the higher dimensional space (with feature map $\phi$), a method is proposed to create a randomized feature map $z$ such that $\phi(x)^T\phi(y)\approx z(x)^Tz(y)$. The approximated feature map in RFF includes sinusoids that are randomly selected from the Fourier transform of the kernel function \cite{RR07}. We use $\gamma=1$ in all our experiments.

We start by investigating the use of Gaussian kernels for a more efficient optimization of the loss function on the training data. Figures \ref{fig:train_loss_mnist} and \ref{fig:train_loss_cifar} show the training loss at different steps respectively on first two classes of MNIST and CIFAR-10 datasets. We observe that Gaussian kernel embeddings with different dimensions yield faster convergence in the optimization on both datasets. MNIST is a simple dataset which gives incredibly high score almost immediately, as shown by the training loss (Figure \ref{fig:train_loss_mnist}) and by the accuracy on the test data (Table \ref{tab:testacc_mnist}). Since we already reach very high accuracy on MNIST without using kernels, different methods yield similar results on this dataset. We observe this behavior in Fashion-MNIST dataset as well. So in this section, we will focus our analysis on the CIFAR-10 dataset. Similar to the setup in \cite{Arora19}, in Figure \ref{fig:train_projections_loss_cifar}, for different methods, we plot the projections of the true class labels on the eigenvectors (i.e., the projections $\{(\mathbf v_i^{T}\mathbf y)^2\}_{i=0}^{n-1}$). For better visualization, we plot the cumulative forms
\begin{equation}
f_i = \sum_{j=0}^i(\mathbf v_j^{T}\mathbf y)^2
\label{eq:overlap_measure}
\end{equation}

\noindent which are normalized such that $\sum_{i=0}^{n-1}(\mathbf v_i^{T}\mathbf y)^2=1$.

The results show that using kernels yields a better alignment of the eigenvectors belonging to the largest eigenvalues and the target labels, leading to faster convergences. In other words, with kernels, we attain larger $(\mathbf v_i^{T}\mathbf y)^2$'s for top eigenvalues. 
%

\begin{table}[ht!]
    \centering
     \caption{Accuracy on MNIST test set (first two classes) after n epochs. We reach a very high accuracy after only a small number of epochs and the test error becomes negligible. }
    \begin{tabular}{l | c c c c c c}
\toprule
         $dim(\mathcal{K})$ & \multicolumn{4}{c}{Steps} \\
        &10 &1000 &50000 &100000 \\
    \midrule
    None &0.998582&        0.999527&             0.999527&        0.999527\\
    500 &       0.996690&        0.999054&               0.999054&       0.998582 \\
     1000 &         0.998109 &         0.999054 &        0.999527 &   0.999054 \\
        3072 &         0.998582 &     0.999527 &         0.999527 &     0.999527 \\
        10000 &         0.998582 &         0.999527 &         0.999527 &    0.999527 \\
        
\bottomrule
    \end{tabular}
    \label{tab:testacc_mnist}
\end{table}

We continue by investigating the generalization performance of the Gaussian kernel method by analyzing the values of equation \eqref{eq:gen_kernel}. Table \ref{tab:yHy} shows this quantity for different settings and kernels respectively on first two classes of MNIST and CIFAR-10 datasets. Please note that for the infinite dimensional case we do not need to perform normalization since for Gaussian kernels the images of input data in the feature space are already normalized\footnote{This comes from the fact that $\mathcal{K}(\bm x, \bm x) = 1$ for Gaussian kernels. For more detail please refer to \cite{scholkopf2002learning}.}. We observe that in both datasets with several kernels we obtain a lower theoretical upper bound on the generalization error compared to the no-kernel case. It is clear that the bound improves as the dimension of the representations increases but also that the generalization bound seems quite sensitive to values of $\gamma$.

\begin{table}[t!]
    \centering
    \caption{Quantification of $  \sqrt{
  \frac{ 2\mathbf y^T(\mathbf H^\infty)^{-1}\mathbf y}{n}}$ (or $  \sqrt{
  \frac{ 2\mathbf y^T(\mathbf H(\mathcal{K})^\infty)^{-1}\mathbf y}{n}}$) for Gaussian kernels  in different experimental settings ($n=12665$ for MNIST and $n=10000$ for CIFAR). For both datasets, most of the Gaussian kernels yield smaller upper bounds on generalization error.}
    \begin{tabular}{l l l  r}
    
    \toprule
    $\gamma$ & Dimension  & MNIST & CIFAR-10 \\
     \midrule
          0.01 & $\infty$ & 0.99 & 7.37 \\
          0.1 & $\infty$ & 0.48 & 3.77\\
          1 & $\infty$ & 0.26 & 1.82\\
          10 & $\infty$ & 1.24   & 0.49\\
          \midrule
          1 & 500 & 0.35  & 3.33\\
          1 & 1000 & 0.32 & 3.20 \\
          1 & 3072 & 0.29 & 3.08\\
          1 & 10000 & 0.27 & 2.98 \\
          \multicolumn{2}{l}{No kernel} &  0.35  &  3.86\\
        \bottomrule
    \end{tabular}
    \label{tab:yHy}
\end{table}

\begin{figure*}[htb!]
    \centering
    \subfigure[Test error on CIFAR-10 data as a function of epoch number.]
    {
        \includegraphics[width=0.45\columnwidth]{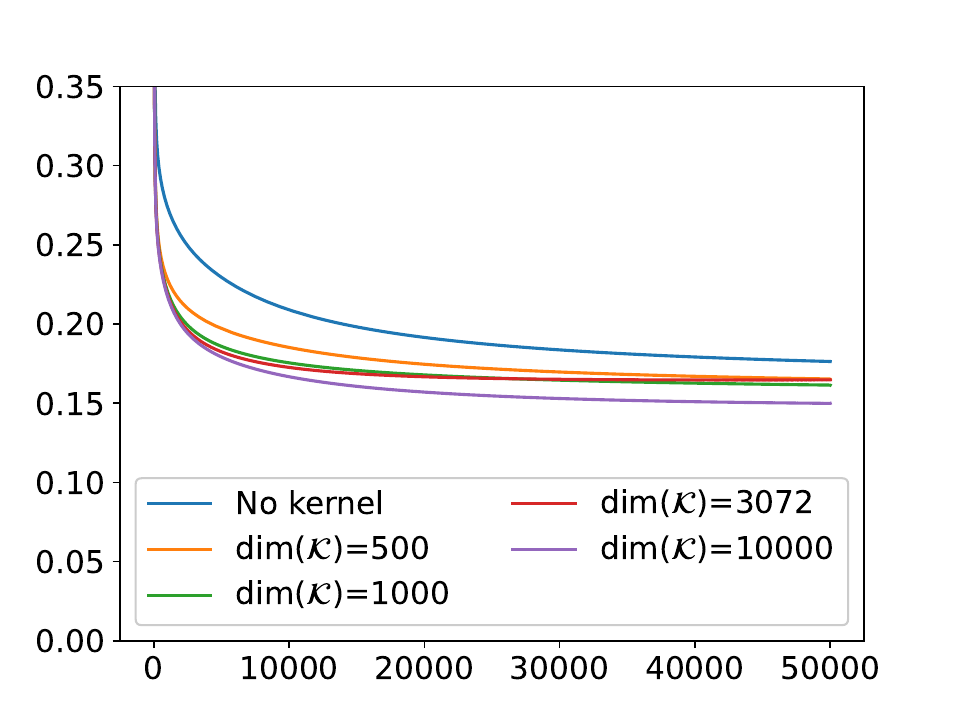}
        \label{fig:testloss_cifar}
    }
    \hspace{3mm}
    \subfigure[Test accuracy on CIFAR-10 data as a function of epoch number.]
    {
        \includegraphics[width=0.45\columnwidth]{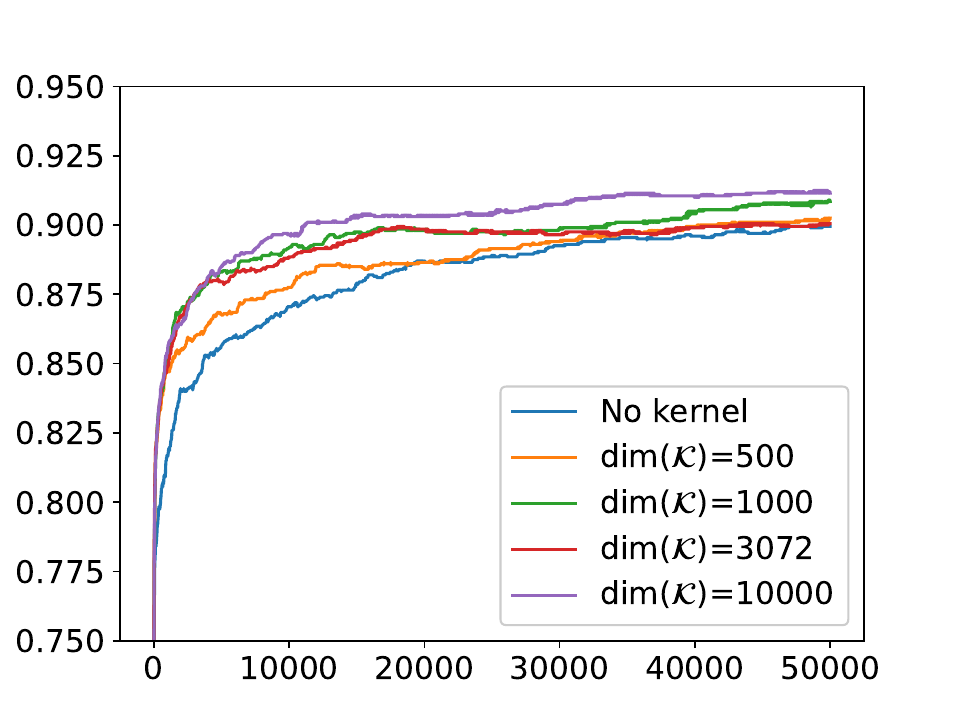}
        \label{fig:testacc_cifar}
    }
    \caption{Experimental test errors and accuracy on the test set at the different steps of the Gradient Descent optimization algorithm for first two classes of CIFAR-10 dataset.
}
    \label{fig:test_accuracy}
\end{figure*}
\begin{figure*}[htb!]
    \centering
    \subfigure[Training loss as a function of epoch number.]
    {
        \includegraphics[width=0.45\textwidth]{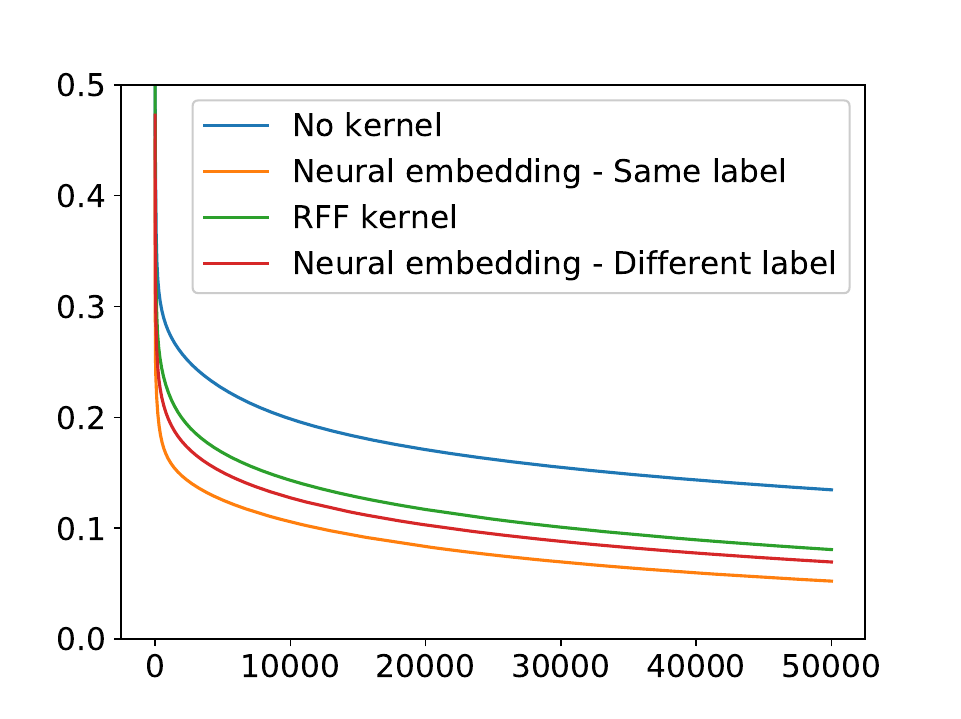}
        \label{fig:trainloss_neural}
    }\hspace{3mm}
    \subfigure[Test loss as a function of epoch number]
    {
        \includegraphics[width=0.45\textwidth]{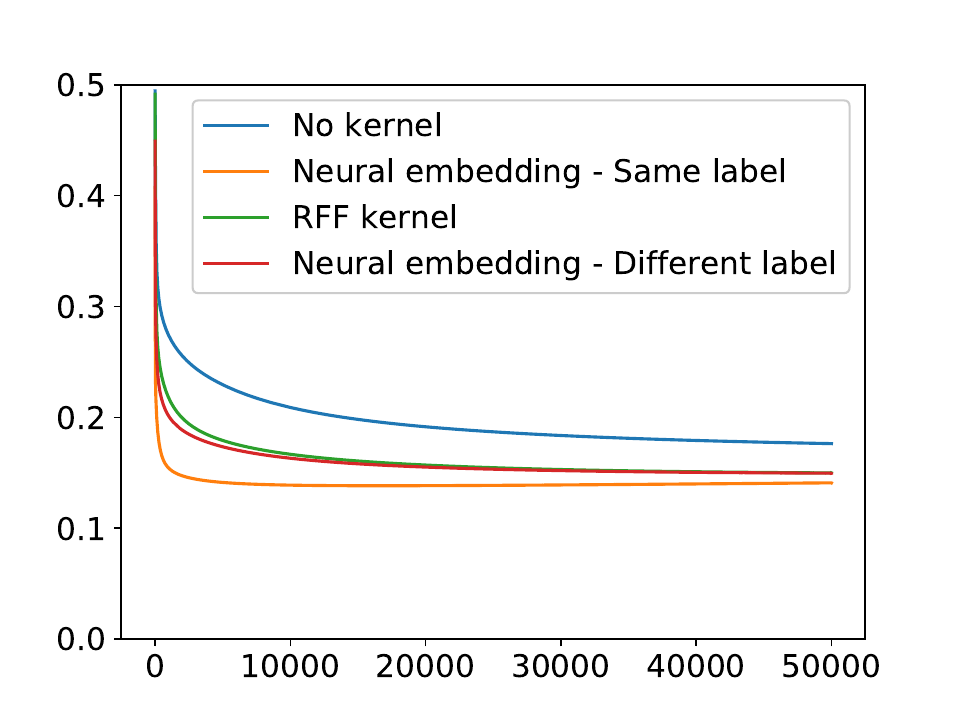}
        \label{fig:testloss_neural}
    }
    \subfigure[Projections]
    {
        \includegraphics[width=0.45\textwidth]{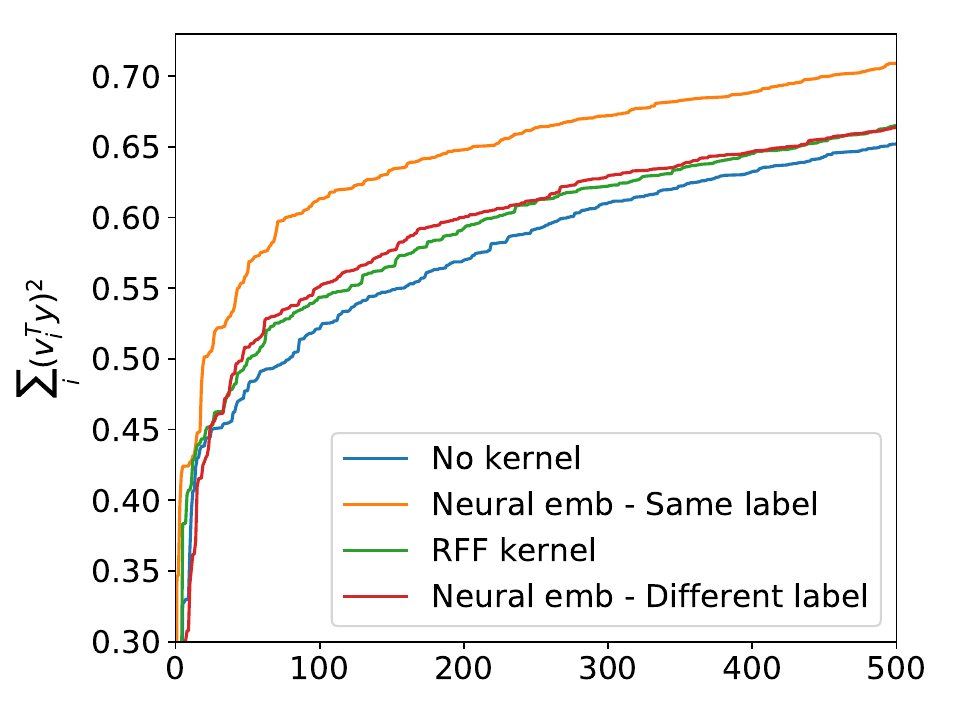}
        \label{fig:train_projections_loss_neural}
    }
    \caption{Experimental train and test errors at the different steps of Gradient Descent as well as eigenvector projections for the first two classes of CIFAR-10 dataset. For the model pre-trained with the same labels, the training loss and projections are calculated based on the unseen subset of training data. We observe that neural embeddings improve the convergence, generalization and the alignment of eigenvector projections.}
    \label{fig:neural_kernel}
\end{figure*}

\begin{figure*}[htb!]
    \centering
    \subfigure[Training loss as a function of epoch number.]
    {
        \includegraphics[width=0.45\textwidth]{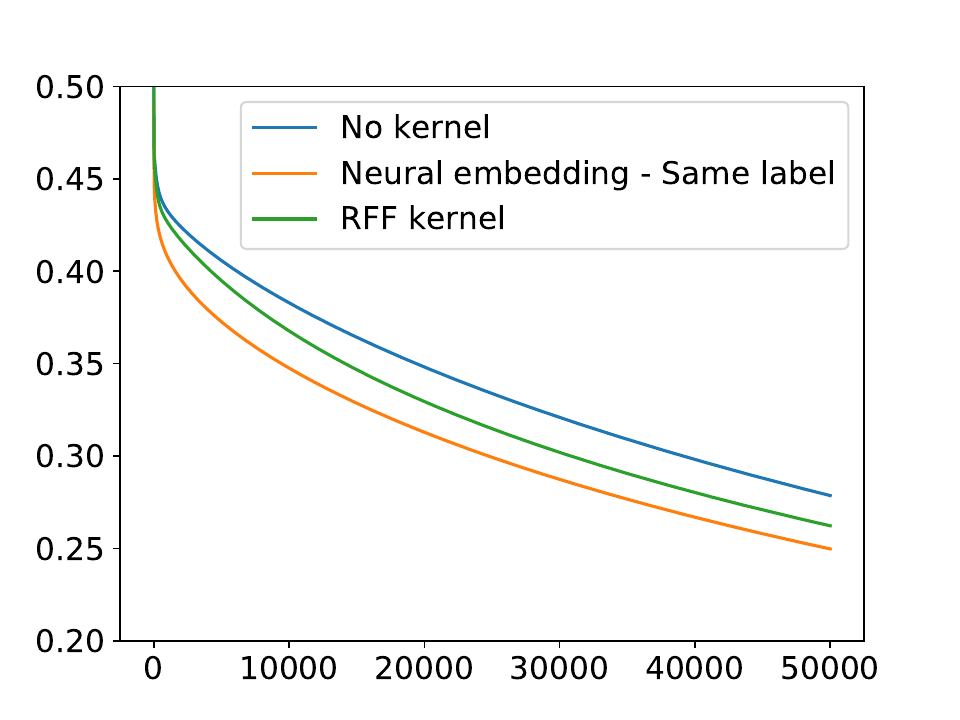}
        \label{fig:trainloss_neural_lsun}
    }\hspace{3mm}
    \subfigure[Test loss as a function of epoch number]
    {
        \includegraphics[width=0.45\textwidth]{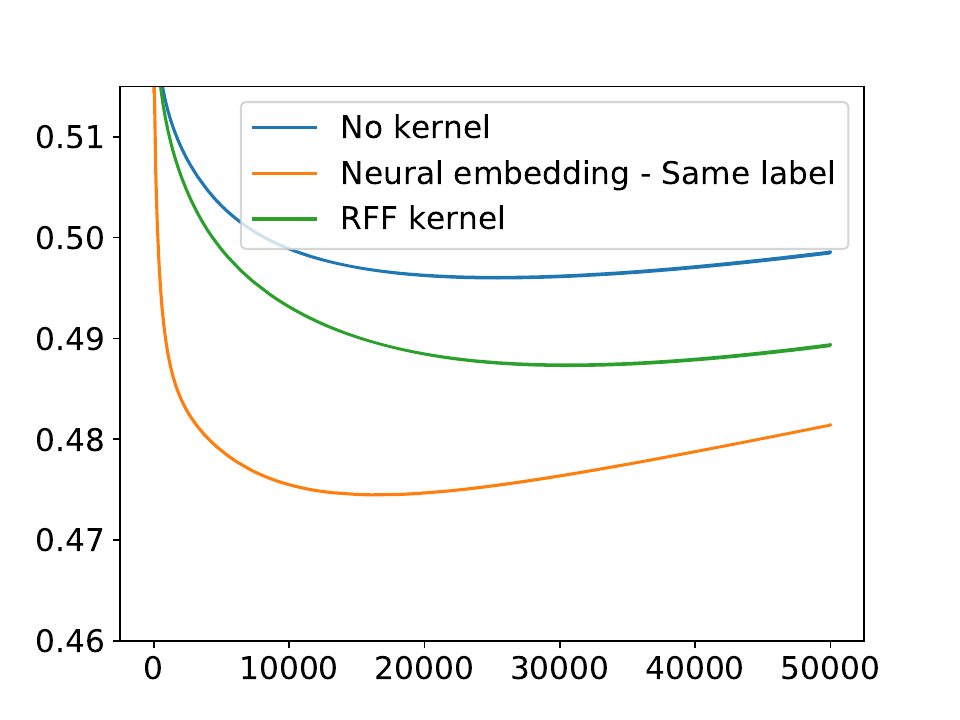}
        \label{fig:testloss_neural_lsun}
    }
    \subfigure[Projections]
    {
        \includegraphics[width=0.45\textwidth]{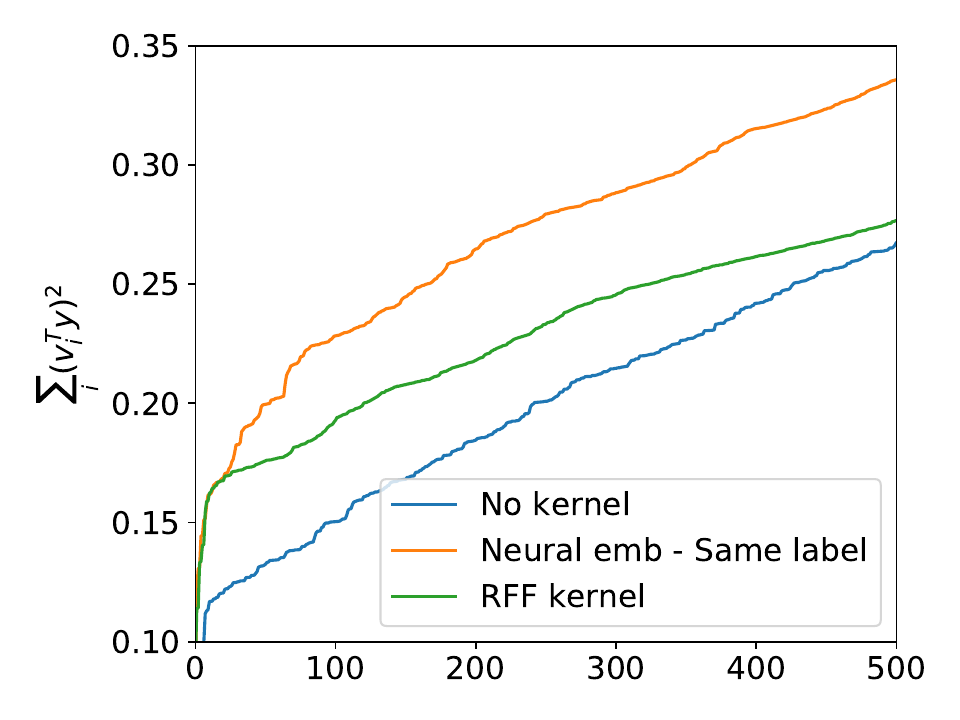}
        \label{fig:train_projections_loss_neural_lsun}
    }
    \caption{Experimental train and test errors at the different steps of Gradient Descent as well as eigenvector projections for the "kitchen" and "living room" classes of LSUN dataset. 
    We observe that neural embeddings improve the convergence, generalization and the alignment of eigenvector projections.}
    \label{fig:neural_kernel_lsun}
\end{figure*}


In addition to the theoretical upper bound, we measure the test error for the studied datasets. Figures \ref{fig:testloss_cifar} and  \ref{fig:testacc_cifar} show respectively the test error and the test accuracy at different steps of the optimization by Gradient Descent for first two classes of  CIFAR-10. Note that in the mentioned figures (and similar figures that follow) we show the test error (loss) and accuracy at each phase of the training. For instance, to compute accuracy we use our current model to predict the labels in a test dataset. We observe that the kernel methods yield significant improvements of both the test error and the accuracy on the test dataset. We observe that the larger the kernel embedding, the larger the improvement. Additionally, we can see a sharper reduction in test error compared to the no-kernel case. This sharp transition (after a small number of steps) comes with a significant improvement in the test accuracy. This observation suggests that early-stopping can be even more efficient when using kernel methods. It is worth noting that these improvements are results of choosing appropriate parameter for the kernels. For instance, a Gaussian kernel with $\gamma=1000$ would not generalize well.
The generalization results shown in Figure \ref{fig:testloss_cifar} are consistent with the measures in Table \ref{tab:yHy}.

\subsection{Neural embedding}
Choosing an appropriate kernel and its parameters can be challenging \cite{vonLuxburg2007}, as also seen in Table \ref{tab:yHy}. Thus, we investigate a data-dependent neural embedding. 
For this purpose, we add a second hidden layer to the neural network with $m=10000$ hidden units and ReLU activation. We pre-train this embedding using two different approaches.  The first layer is then kept fix as an embedding where the rest of the network is reinitialized and trained. The first approach is to split the training data in half. We use the first subset to pre-train this three-layer network and the second subset to use for our optimization experiments. In this approach we double $\eta$ to keep the step length the same. The other approach is to use data from a different domain for pre-training. For instance, we use the last two classes of the CIFAR-10 dataset for pre-training the embedding. We compare our results with not using any kernel and with using an RFF kernel with embedding of size 10000. In Figure \ref{fig:neural_kernel} we show results for the two mentioned settings on first two classes of CIFAR-10: i. using a subset of the training data for pre-training (same label), ii. using data from other classes for pre-training (different label). Also in Figure \ref{fig:neural_kernel_lsun} we show the results of experiments on "kitchen" and "living room" classes of LSUN with the same label setting. Figures \ref{fig:neural_kernel} and \ref{fig:neural_kernel_lsun} show the average of 5 different random initializations. We observe very insignificant standard deviation in different stages of training. For instance, both Gaussian and neural kernels show standard deviation of order $10^{-4}$ for training loss when using the LSUN dataset in different steps of training.

\subsubsection{Optimization}
Figure \ref{fig:trainloss_neural} shows the training loss for the first two classes CIFAR-10 dataset. We observe the neural embeddings achieve faster convergence compared to the previous methods. We report the training loss for neural embedding (same label) on the second (unused) subset of the data, whereas in the other cases we report the results on the full training data. If we use only the second subset for the other methods, we observe very consistent results to Figure \ref{fig:neural_kernel}. 
Figure \ref{fig:train_projections_loss_neural} demonstrates eigenvector projections on the target labels. This shows that both variants of neural embeddings improve alignment of the labels to eigenvectors corresponding to larger eigenvalues (compared to the best RFF kernel). While the effect is unsurprisingly larger when pre-training on the same labels, it is still significantly better when pre-trained on other labels. Also in Figures \ref{fig:trainloss_neural_lsun} and \ref{fig:train_projections_loss_neural_lsun}, we see similar results for the LSUN dataset, and the neural embedding outperforms RFF kernel in terms of training loss. The alignment of labels and eigenvectors is improved as well.

\subsubsection{Generalization}
In Figures \ref{fig:testloss_neural} and \ref{fig:testloss_neural_lsun} we report the test error on the first two classes CIFAR-10 and "kitchen" and "living room" classes of LSUN respectively. This shows that the neural embeddings perform at least comparable with the best studied RFF kernel. If the  pre-training is done on the same labels we obtain a clear improvement, even if the actual training is only done on a dataset with half the size.

Finally, it is worth mentioning that similar results are obtained when the pre-training process is stopped earlier (i.e., 5,000 epochs).


 \subsection{Embeddings with other kernels}

So far, we have used two embedding methods: i) a fixed  embedding according to Gaussian kernels, and ii)  the embeddings induced by deep neural networks. In this section we analyze the use of kernels designed for our task of interest. In particular, we study two types of kernels: i) the arc-cosine kernel proposed in \cite{CS09} which represents neural network layers. Note that, as we will discuss, computing this kernel is significantly faster than training a neural network layer. Thus, it can be used for the same purpose, i.e., an embedding induced by a neural network, but in a more efficient way. ii) The multiple kernel learning method proposed in \cite{CortesMohri2012}. Given the training data and the corresponding labels, this method computes the optimal feasible kernel, independent of any classification model or algorithm. We call the resulting kernel optimal since an optimization function is defined and the final kernel is found based on the solution from the optimization problem (see section \ref{sec:modelfree}). 

In both cases, we first approximate the kernel embedding using the Nystr\"{o}m method \cite{Nystroem}, and then feed the approximated kernel embedding to the two-layer neural network model, instead of the original data features.


\begin{figure*}[ht!]
    \centering
    \subfigure[Training loss as a function of epoch number.]
    {
        \includegraphics[width=0.30\textwidth]{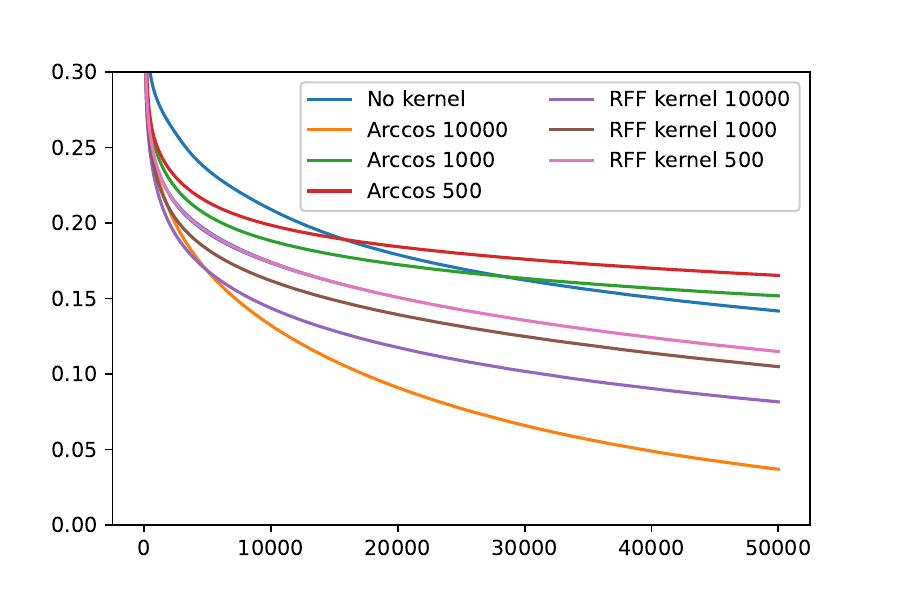}
        \label{fig:trainloss_arccosine_cifar01}
    }
    \hspace{1mm}
    \subfigure[Test loss as a function of epoch number]
    {
        \includegraphics[width=0.30\textwidth]{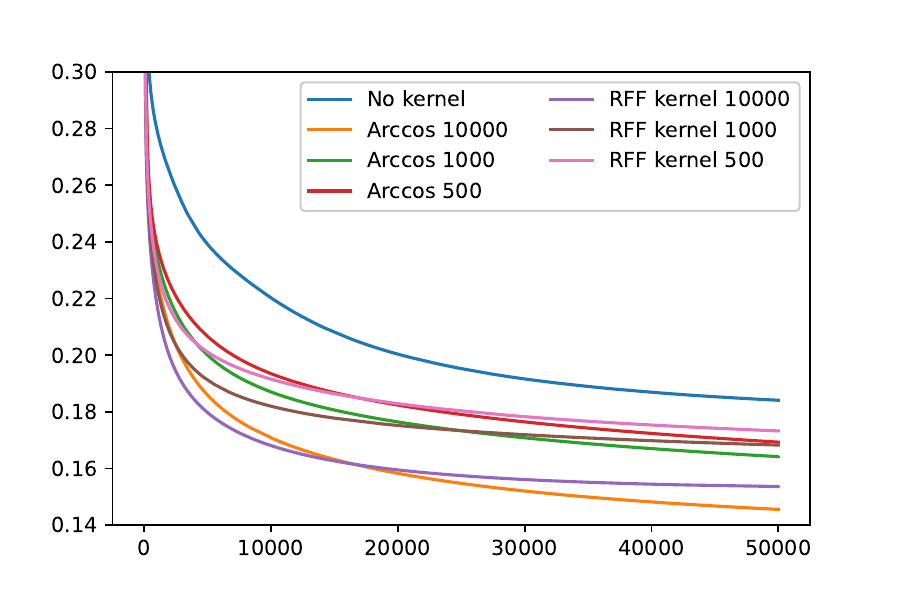}
        \label{fig:testloss_arccosine_cifar01}
    }
    \subfigure[Projections]
    {
        \includegraphics[height=0.19\textwidth,width=0.30\textwidth]{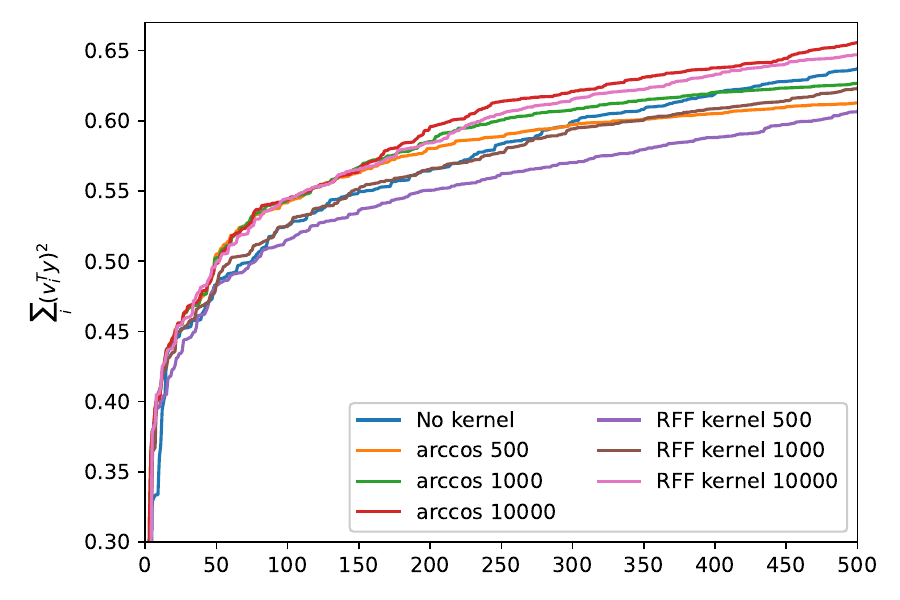}
        \label{fig:train_projections_loss_arccosine_cifar01}
    }
    \subfigure[Training loss as a function of epoch number.]
    {
        \includegraphics[width=0.30\textwidth]{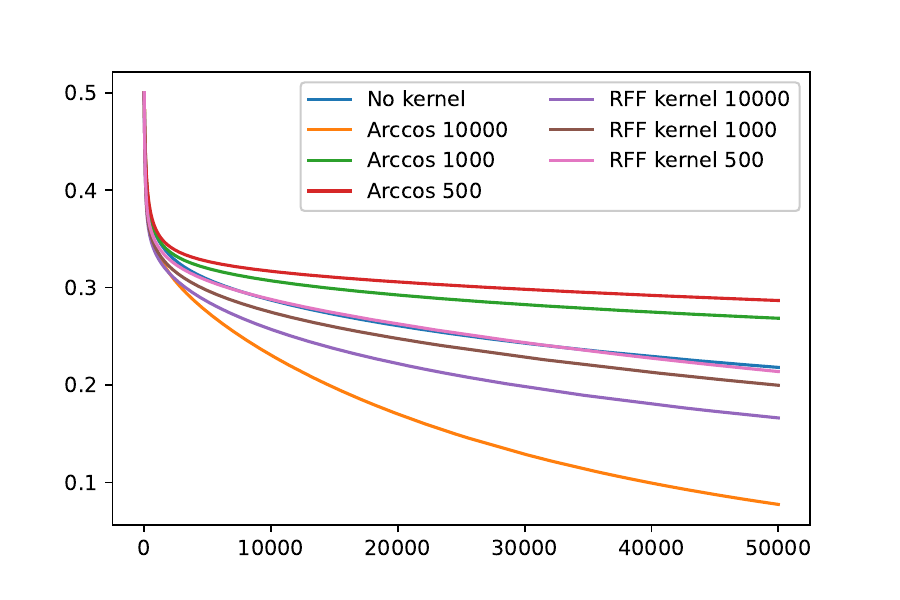}
        \label{fig:trainloss_arccosine_cifar23}
    }
    \hspace{1mm}
    \subfigure[Test loss as a function of epoch number]
    {
        \includegraphics[width=0.30\textwidth]{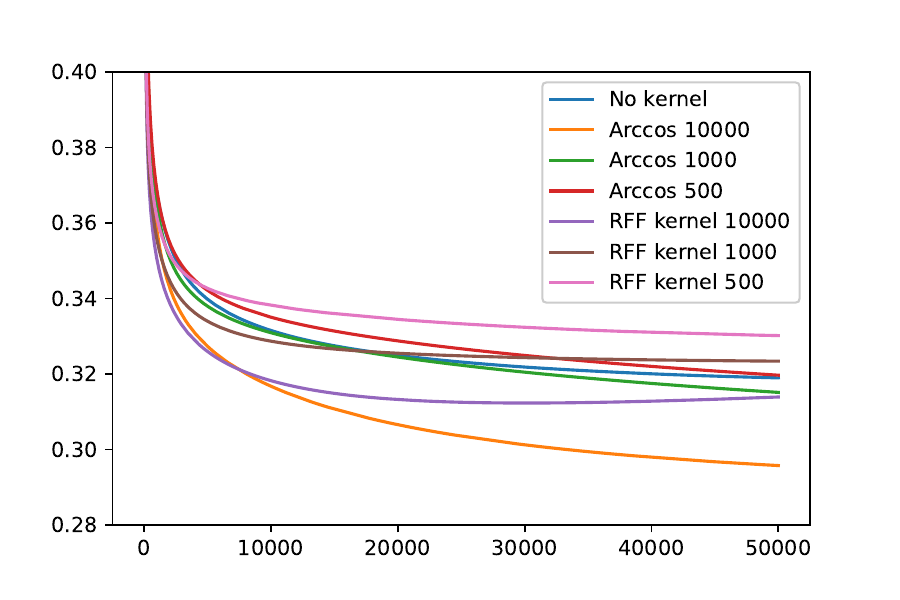}
        
        \label{fig:testloss_arccosine_cifar23}
    }
    \subfigure[Projections]
    {
        \includegraphics[height=0.19\textwidth,width=0.30\textwidth]{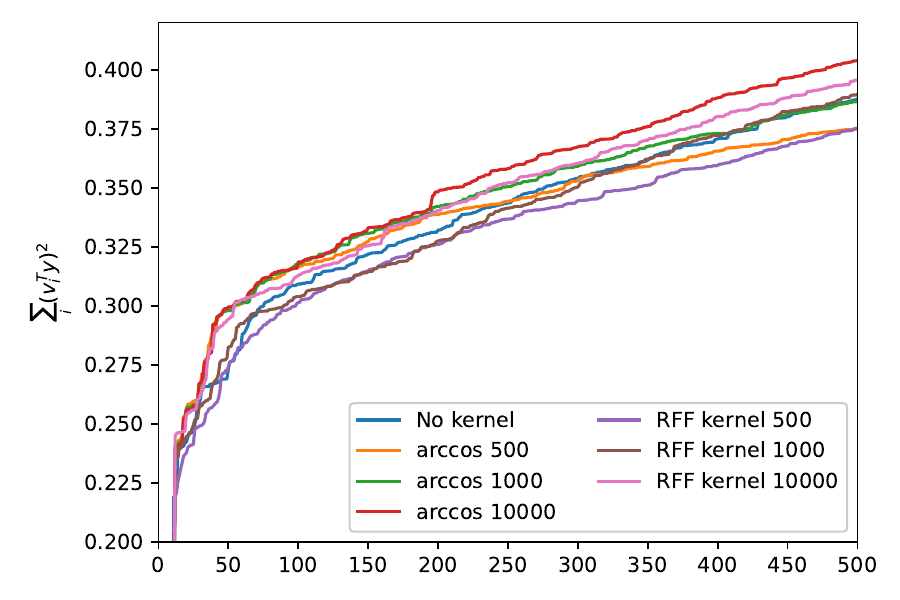}
        \label{fig:train_projections_loss_arccosine_cifar23}
    }
    \subfigure[Training loss as a function of epoch number.]
    {
        \includegraphics[width=0.30\textwidth]{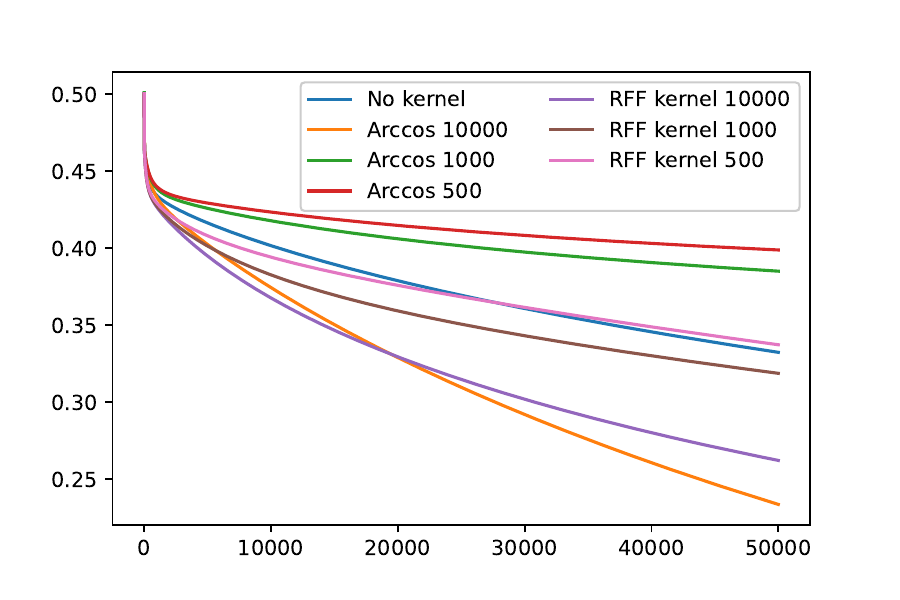}
        \label{fig:trainloss_arccosine_lsun}
    }
    \hspace{1mm}
    \subfigure[Test loss as a function of epoch number]
    {
        \includegraphics[width=0.30\textwidth]{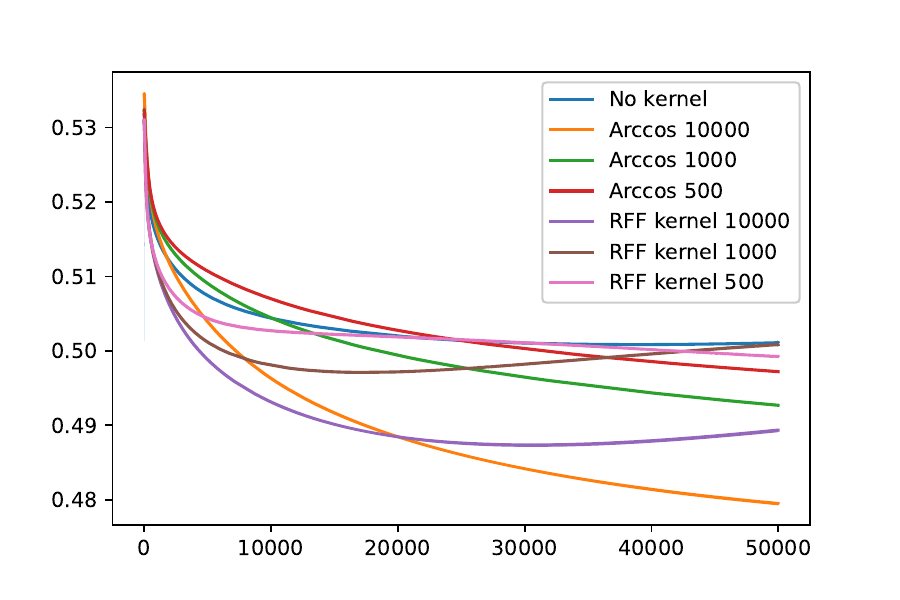}
        \label{fig:testloss_arccosine_lsun}
    }
    \subfigure[Projections]
    {
        \includegraphics[height=0.19\textwidth,width=0.30\textwidth]{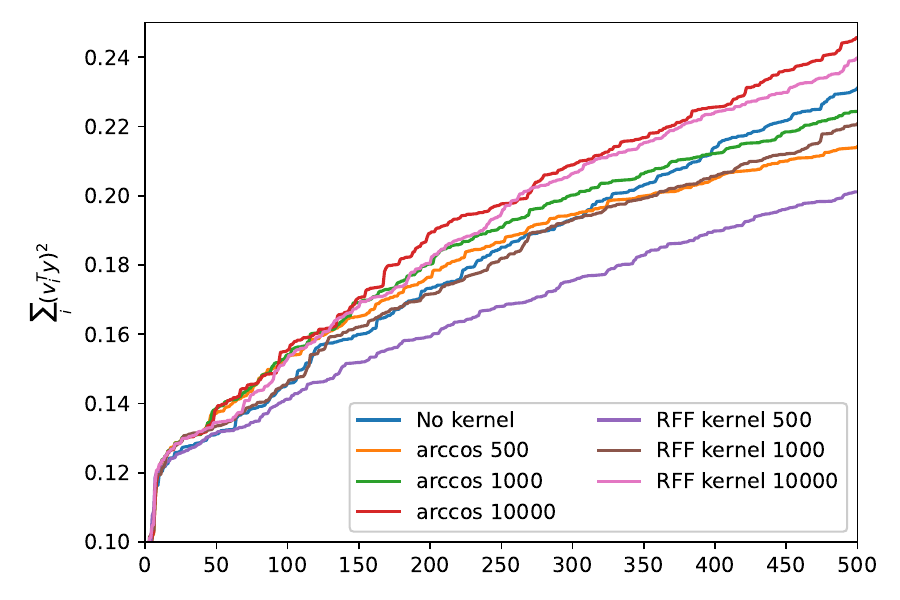}
        \label{fig:train_projections_loss_arccosine_lsun}
    }
    
    \caption{The results of the embeddings induced by the arc-cosine kernel on first two classes of CIFAR-10 (a-c), second and third classes of CIFAR-10 (d-f) and ”kitchen” and ”living room” classes of LSUN (g-i) with different dimensionality of parameterizations. This kernel  performs better than the Gaussian kernel for sufficiently large number of components and the case where we use the original features as in \cite{Arora19}. The overlap measure in Eq. \ref{eq:overlap_measure} is verified for this case as well.}
    \label{fig:arccosine_kernel_aggregated}
\end{figure*}

\subsubsection{Arc-cosine kernel}
The arc-cosine kernel \cite{CS09} mimics the computations in a neural network within an infinite dimensional feature space, where the kernel function is equivalent to the inner product of the feature vectors from a neural network. Then, the Nystr\"{o}m method with different number of components can be used to approximate the kernel embeddings. 
 
The $n$-th order \emph{arc--cosine} kernel between two inputs $x$ and $y$ is computed by
 
\begin{equation}
 k_n(x,y) = \frac{1}{\pi} \|x\|^n \|y\|^n J_n(\theta).
\end{equation}

Where the function $J_n$ and $\theta$ are given by

\begin{equation}
 J_n(\theta) = (-1)^n(\sin\theta)^{2n+1}{\left(\frac{1}{\sin\theta}\frac{\partial}{\partial\theta}\right)}^n\left(\frac{\pi-\theta}{\sin\theta}\right)
\end{equation}

\begin{equation}
\theta = \cos^{-1}\left(\frac{x.y}{\|x\| \|y\|}\right)
\end{equation}

In \cite{CS09} it is shown that the computations in single-layer threshold networks is closely related to this kernel. Specifically, the kernel is equivalent to the inner product of embeddings induced by a single-layer threshold network.
Moreover, \cite{CS09} suggests a family of kernels mimicking multi-layer networks based on the base kernel. In our experiments we use the base kernel with $n=0$\footnote{The reason behind this particular choice is that in \cite{CS09} it is mentioned that best results are achieved by $n=0$ or $1$.}. It is worth mentioning that earlier, but with different activation functions, these kernels arose as covariance functions of limiting Gaussian processes of wide neural networks \cite{BayesNN}. Also in \cite{covNN} closed form kernels for different activation functions are derived. Moreover, in \cite{TKG18} the kernel for Leaky ReLU activations is calculated.

Figures \ref{fig:trainloss_arccosine_cifar01}-\ref{fig:train_projections_loss_arccosine_cifar01}, \ref{fig:trainloss_arccosine_cifar23}-\ref{fig:train_projections_loss_arccosine_cifar23} and \ref{fig:trainloss_arccosine_lsun}-\ref{fig:train_projections_loss_arccosine_lsun} show the results of the arc-cosine kernel embeddings on first two classes of CIFAR-10, second and third classes of CIFAR-10, and "kitchen" and "living room" classes of LSUN respectively. We conducted these experiments with 5 different random initializations and these figures illustrate the average results. We observe that arc-cosine kernel (with a sufficiently large number of components) outperforms the Gaussian kernels as well as the no-kernel case in all of our experiments. The standard deviation of our results in different stages of training was very insignificant. For example, for the arc-cosine and Gaussian kernel embeddings of size 10000, the standard deviations of training loss are in the order of $10^{-4}$. Similar to the Gaussian kernel increasing the number of components in arc-cosine kernel embeddings improves the estimations, and hence, leads to a faster optimization and a lower test error. Moreover, the overlap parameters are also better for  arc-cosine kernel with a large number of compoenents (i.e., arccos 10000). These results show that assuming the arccosine kernel does approximate neural embeddings, then the results of \cite{Arora19} extend to multilayer networks.

 Notice that Neural Tangent Kernel (NTK) \cite{JFH18} is another  approach to compute a kernel representing a neural network. However, the applicability of this method is confined by its very inefficient computational runtime. By inefficiency we mean it is not a good choice to use NTK together with Nystr\"{o}m method.\footnote{In our experiments on even a significantly smaller dataset (i.e., with only $1000$ images) it took about 24 hours to run the algorithm on GPU.}


\begin{figure*}
    \centering
    \subfigure[Training loss as a function of epoch number.]
    {
        \includegraphics[width=0.30\textwidth]{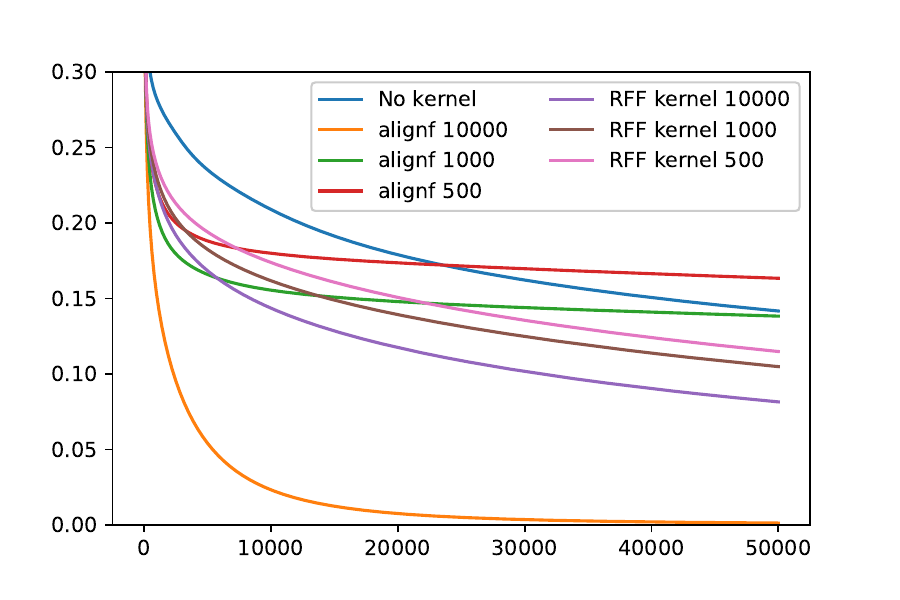}
        \label{fig:trainloss_alignf_cifar01}
    }
    \hspace{1mm}
    \subfigure[Test loss as a function of epoch number.]
    {
        \includegraphics[width=0.30\textwidth]{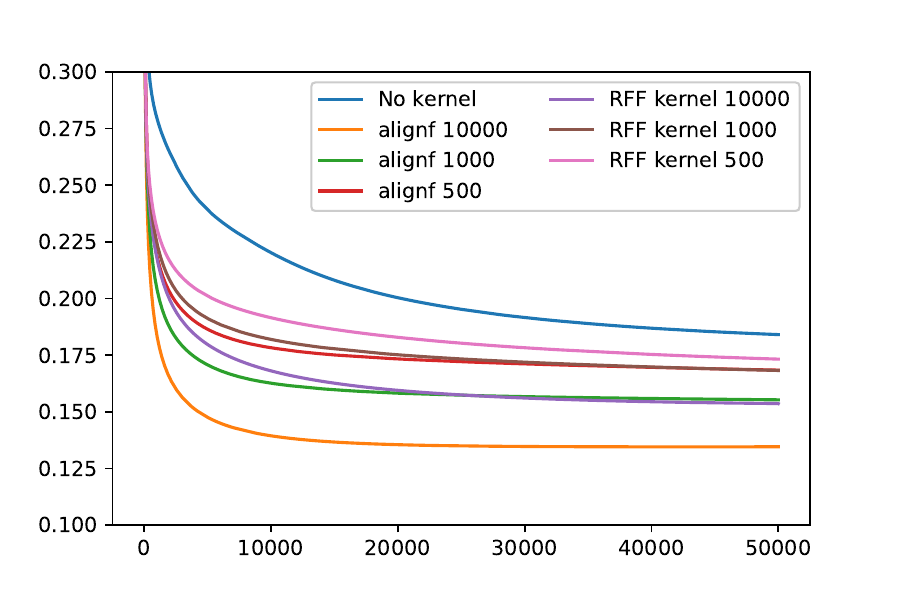}
        \label{fig:testloss_alignf_cifar01}
    }
    \subfigure[Projections]
    {
        \includegraphics[height=0.19\textwidth,width=0.30\textwidth]{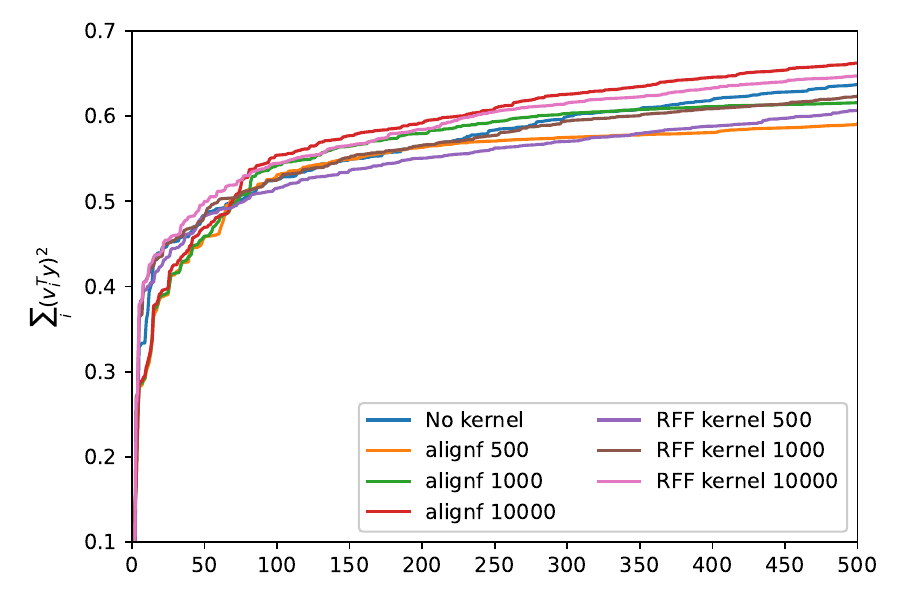}
        \label{fig:train_projections_loss_alignf_cifar01}
    }
    \subfigure[Training loss as a function of epoch number.]
    {
        \includegraphics[width=0.30\textwidth]{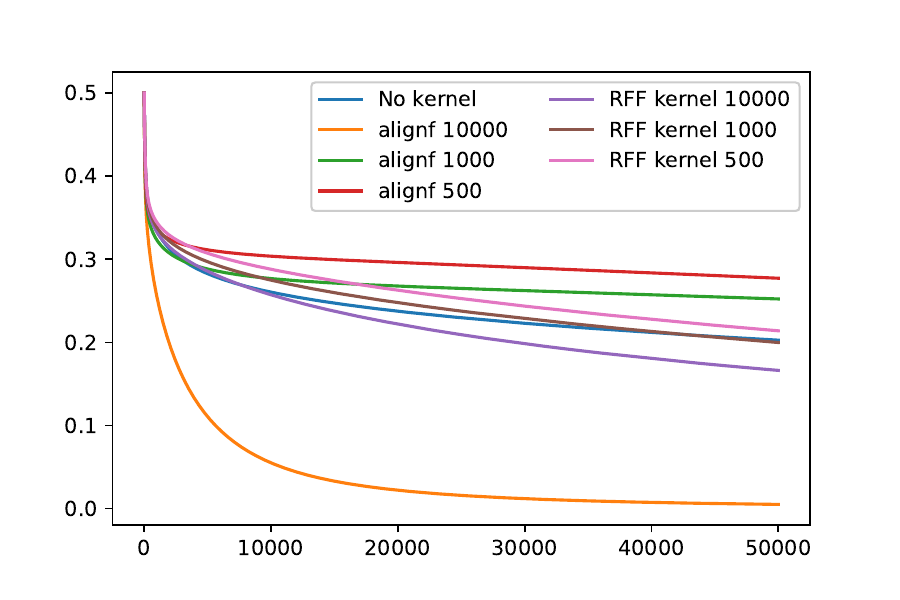}
        \label{fig:trainloss_alignf_cifar23}
    }
    \hspace{1mm}
    \subfigure[Test loss as a function of epoch number.]
    {
        \includegraphics[width=0.30\textwidth]{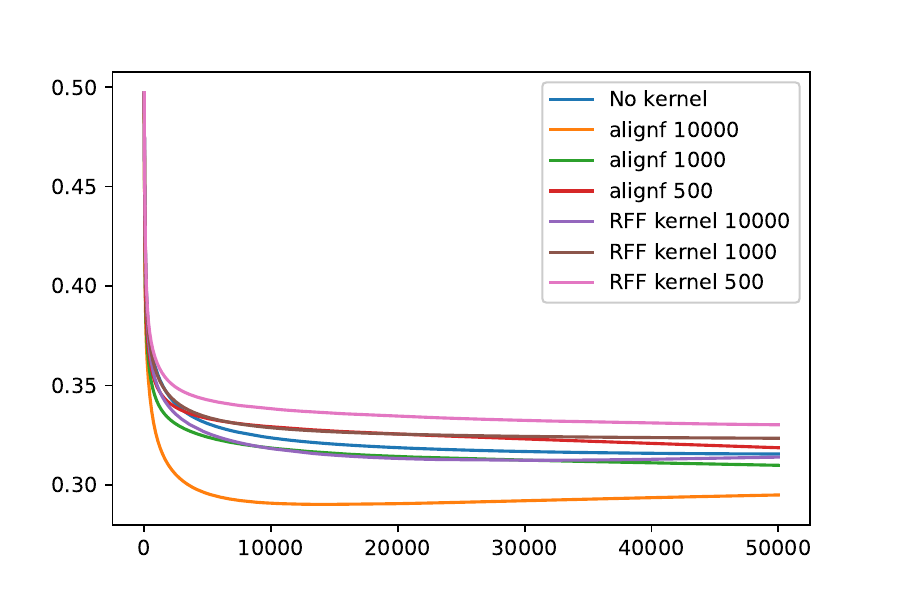}
        \label{fig:testloss_alignf_cifar23}
    }
    \subfigure[Projections]
    {
        \includegraphics[height=0.19\textwidth,width=0.30\textwidth]{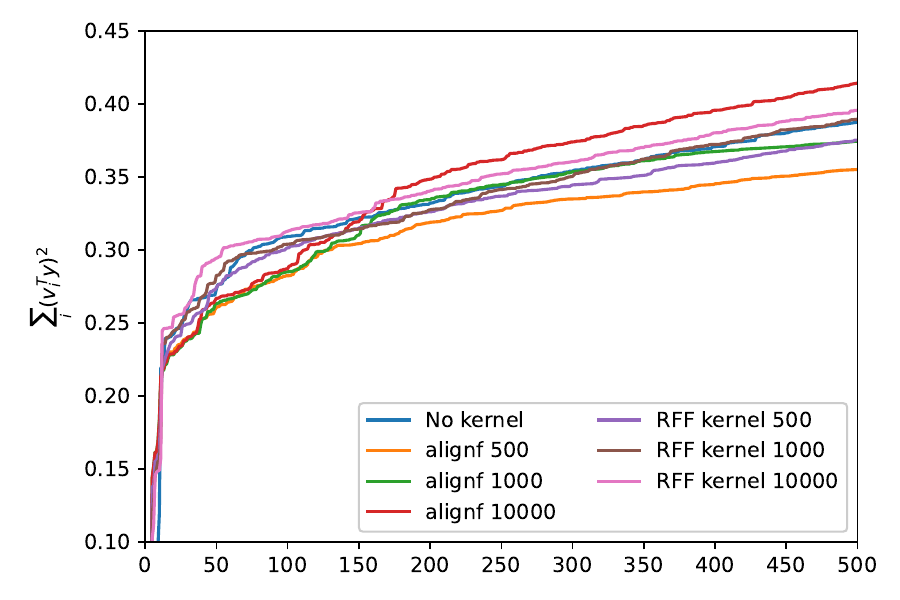}
        \label{fig:train_projections_loss_alignf_cifar23}
    }
    \subfigure[Training loss as a function of epoch number.]
    {
        \includegraphics[width=0.30\textwidth]{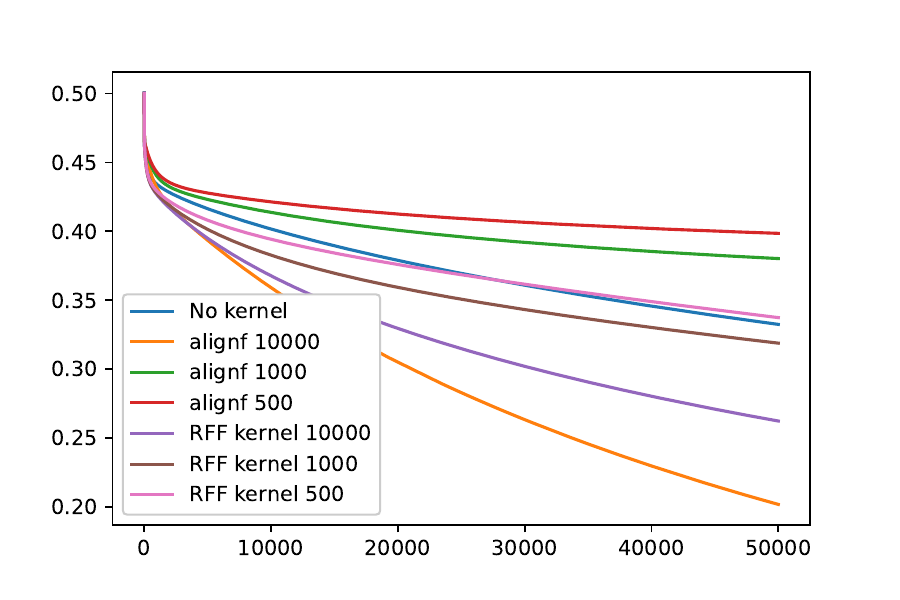}
        \label{fig:trainloss_alignf_lsun}
    }
    \hspace{1mm}
    \subfigure[Test loss as a function of epoch number.]
    {
        \includegraphics[width=0.30\textwidth]{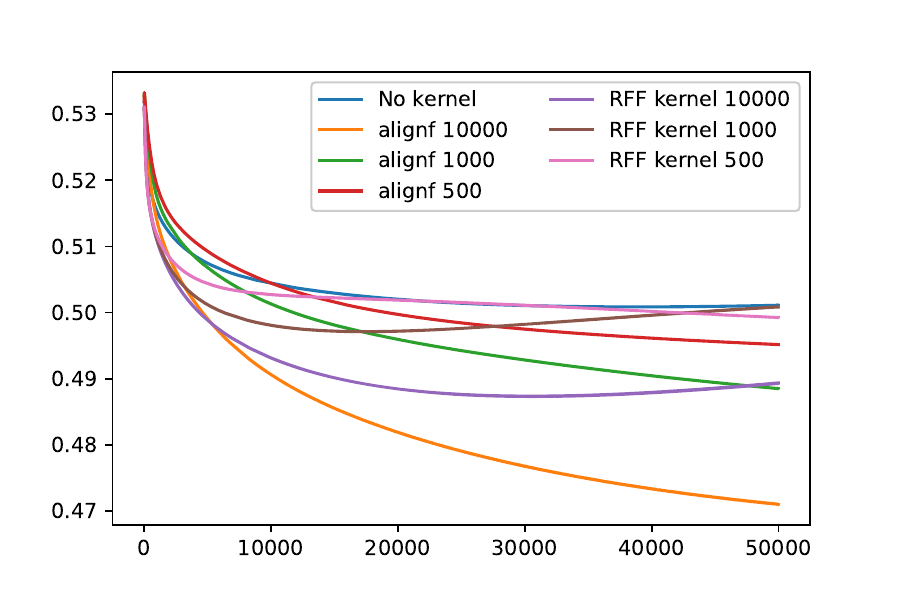}
        \label{fig:testloss_alignf_lsun}
    }
    \subfigure[Projections]
    {
        \includegraphics[height=0.19\textwidth,width=0.30\textwidth]{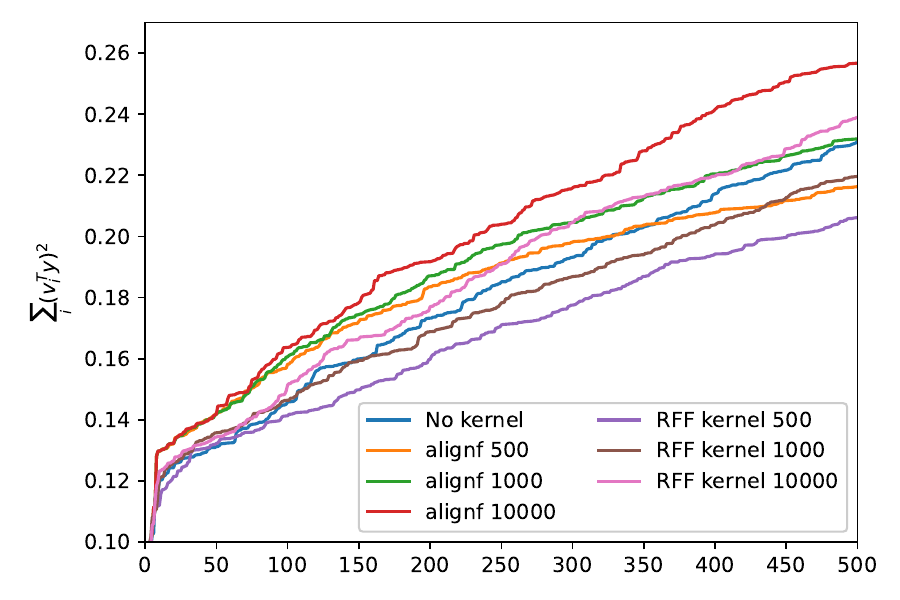}
        \label{fig:train_projections_loss_alignf_lsun}
    }
    \caption{The results of the embeddings induced by the \emph{alignf} kernel on first two classes of CIFAR-10 (a-c), second and third classes of CIFAR-10 (d-f), and ”kitchen” and ”living room” classes of LSUN (g-i) with different dimensionality of parameterizations. This kernel is optimally learned w.r.t. the data and significantly performs better than the arc-cos kernel, the Gaussian kernel and the no-kernel case. We again observe the consistency of a fast optimization, a low test error and a large overlap measure defined in Eq. \ref{eq:overlap_measure}.}
    \label{fig:alignf_kernel_aggregated}
\end{figure*}

\subsubsection{Multiple kernel learning}
\label{sec:modelfree}
Finally, we investigate learning an appropriate kernel, independent of the model or the algorithm to be used for training and classification. For this purpose, we use the method proposed in \cite{CortesMohri2012} that suggests an algorithm to learn a new kernel from a group of kernels based on a similarity measure between the kernels, namely centered alignment. Then, the problem of learning a kernel with a maximum  alignment between the data and the labels is formulated as a quadratic programming (QP) problem. The respective algorithm is called \emph{alignf} \cite{CortesMohri2012}. This multiple kernel learning algorithm does not involve parameter selection and as mentioned it is computed solely based on the true class labels. 
The kernel learning algorithm works based on \emph{centered kernel alignment}. If $K_1$ and $K_2$ are two kernel matrices, the centered alignment between them can be  computed by the following \cite{Gonen2011}.

\begin{equation}
 CA(K_1,K_2) = \frac{\langle K_1^c,K_2^c\rangle_F}{\sqrt{\langle K_1^c,K_1^c\rangle_F\langle K_2^c,K_2^c\rangle_F}} ,
\end{equation}

where $K^c$ is the centered version of the kernel matrix $K$. To find the optimal combination of the kernels (i.e. a weighted combination of some base kernels), \cite{CortesMohri2012} suggests the objective function to be centered alignment between the combination of the kernels and $yy^T$, where $y$ is the labels vector. By restricting the weights to be non-negative, a QP can be obtained as

\begin{equation}
 minimize\hspace{4px} v^TMv - 2v^Ta \hspace{4px}
 w.r.t.\hspace{2px} v \in R_+^P
\end{equation}

 $P$ is the number of the base kernels and $M_{kl} = \langle K_k^c,K_l^c\rangle_F$ for $k,l \in [1,P]$, and finally $a$ is a vector wherein $a_i = \langle K_i^c,yy^T\rangle_F$ for $i \in [1,P]$. If $v^*$ is the solution of the QP, then the vector of kernel weights is given by \cite{CortesMohri2012,Gonen2011}

\begin{equation}
\mu^* = v^*/\|v^*\|
\end{equation}


\begin{figure*}
    \centering
    \subfigure[Training loss as a function of epoch number.]
    {
        \includegraphics[width=0.30\textwidth]{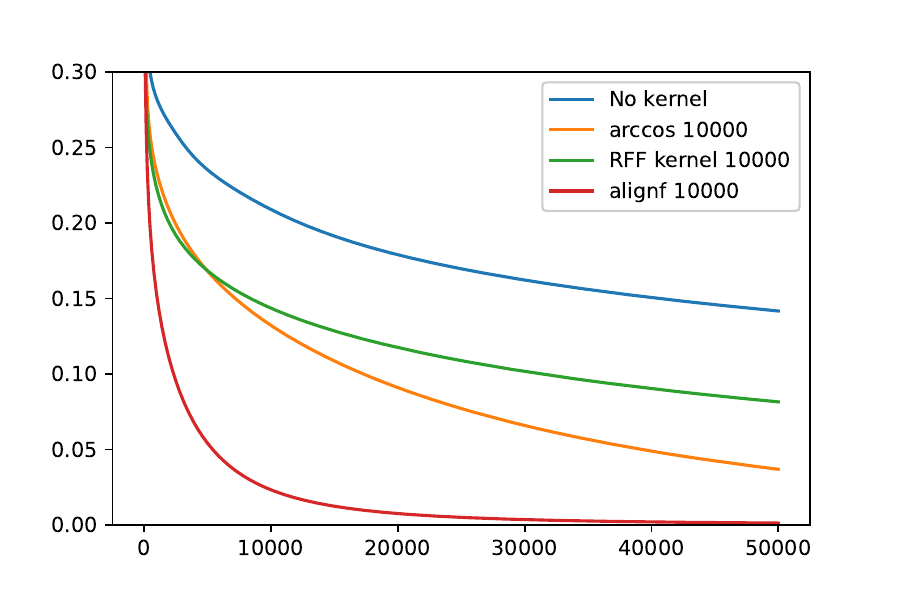}
        \label{fig:trainloss_bestkernel_cifar01}
    }
    \hspace{1mm}
    \subfigure[Test loss as a function of epoch number.]
    {
        \includegraphics[width=0.30\textwidth]{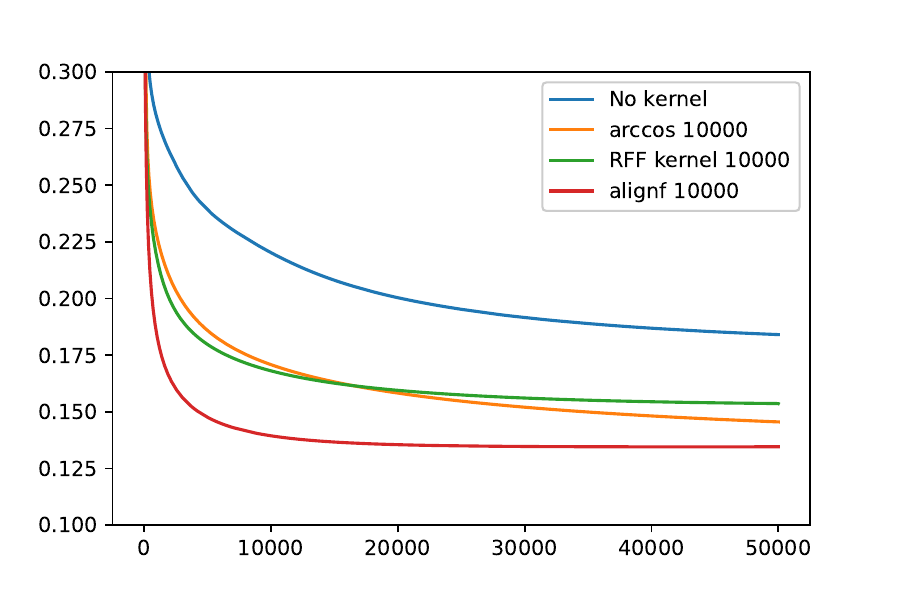}
        \label{fig:testloss_bestkernel_cifar01}
    }
    \subfigure[Projections]
    {
        \includegraphics[height=0.19\textwidth,width=0.30\textwidth]{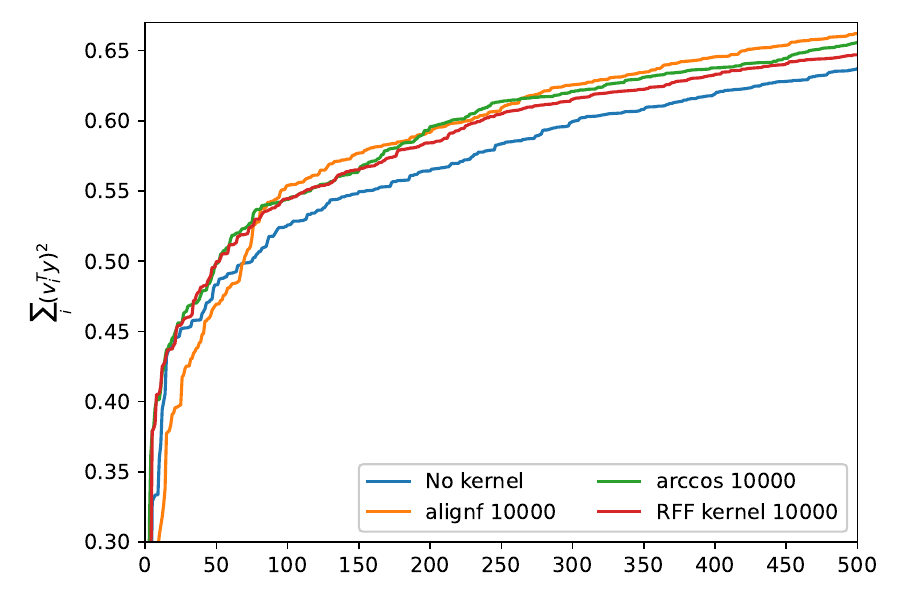}
        \label{fig:train_projections_bestkernel_cifar01}
    }
    \subfigure[Training loss as a function of epoch number.]
    {
        \includegraphics[width=0.30\textwidth]{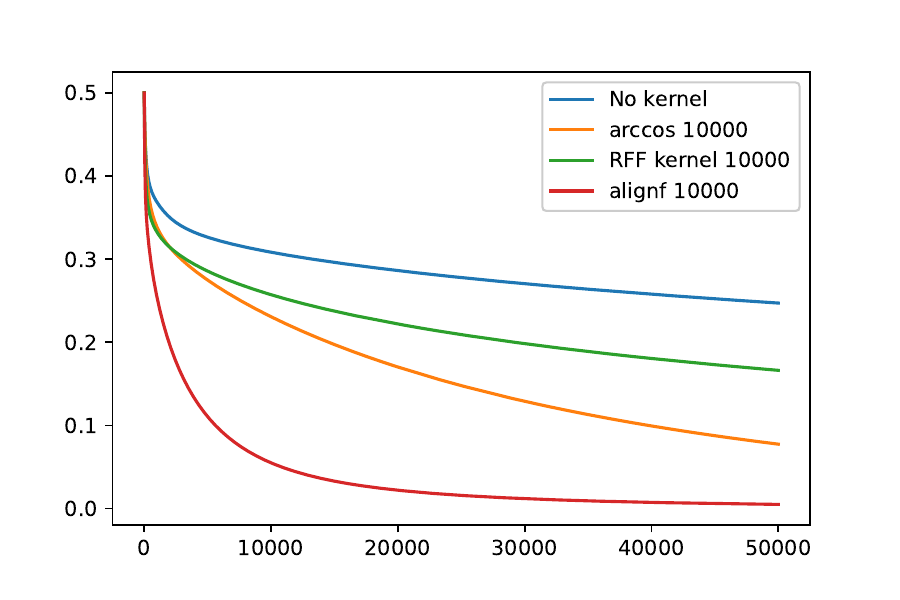}
        \label{fig:trainloss_bestkernel_cifar23}
    }
    \hspace{1mm}
    \subfigure[Test loss as a function of epoch number.]
    {
        \includegraphics[width=0.30\textwidth]{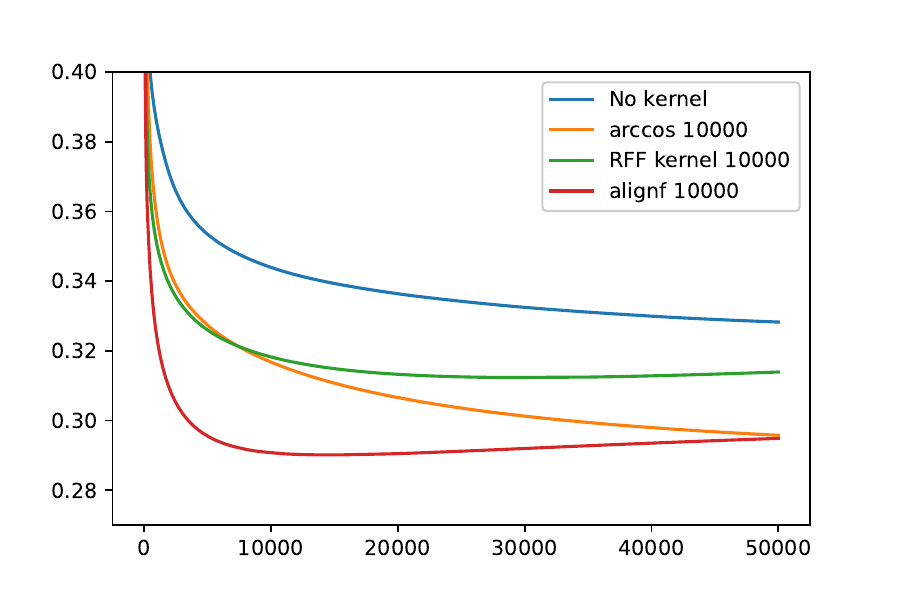}
        \label{fig:testloss_bestkernel_cifar23}
    }
    \subfigure[Projections]
    {
        \includegraphics[height=0.19\textwidth,width=0.30\textwidth]{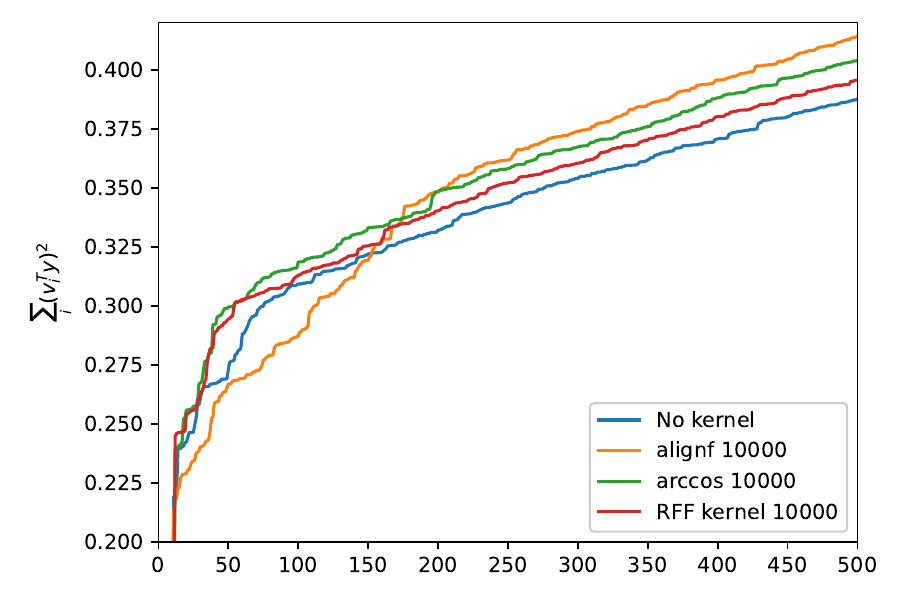}
        \label{fig:train_projections_bestkernel_cifar23}
    }
    \subfigure[Training loss as a function of epoch number.]
    {
        \includegraphics[width=0.30\textwidth]{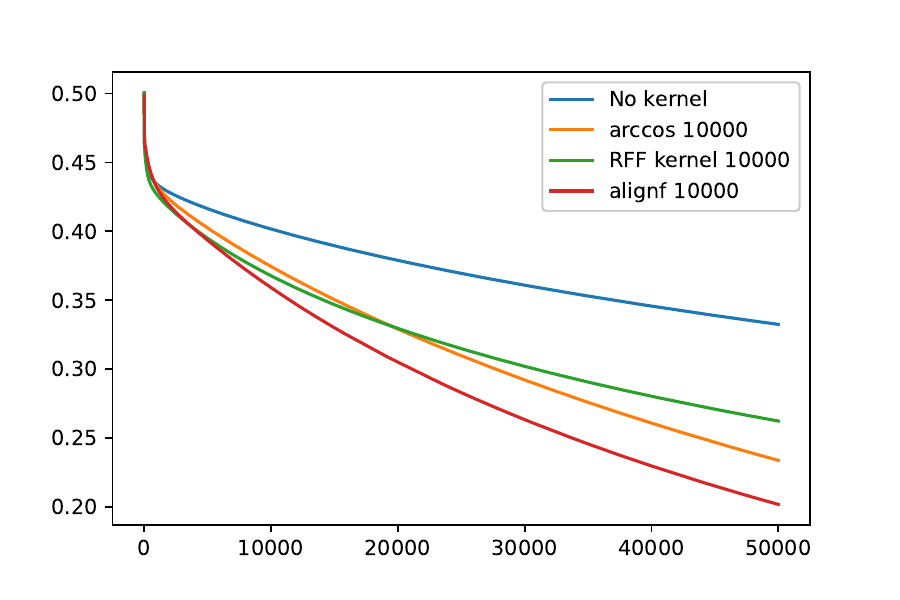}
        \label{fig:trainloss_bestkernel_lsun}
    }
    \hspace{1mm}
    \subfigure[Test loss as a function of epoch number.]
    {
        \includegraphics[width=0.30\textwidth]{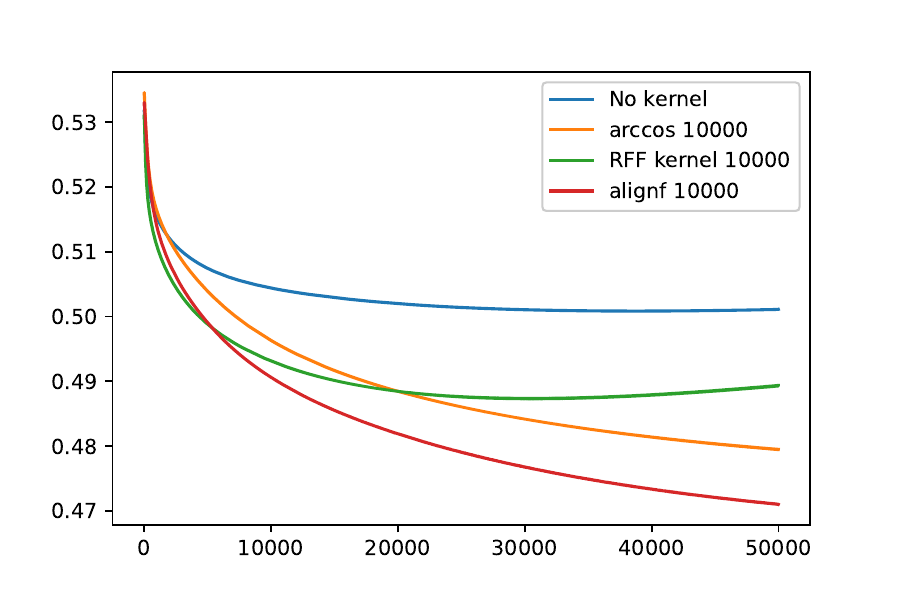}
        \label{fig:testloss_bestkernel_lsun}
    }
    \subfigure[Projections]
    {
        \includegraphics[height=0.19\textwidth,width=0.30\textwidth]{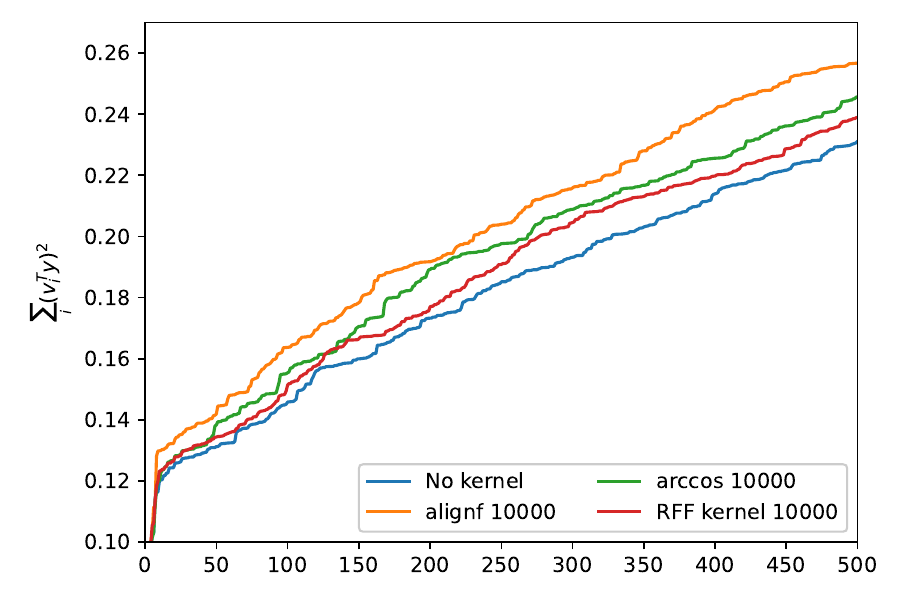}
        \label{fig:train_projections_bestkernel_lsun}
    }
    \caption{The consistency of fast optimization, low test error and high overlap between the labels vector and the top eigenvectors using the first two classes of CIFAR-10 (a-c), second and third classes of CIFAR-10 (d-f), and ”kitchen” and ”living room” classes of LSUN (g-i) as the dataset. We see a consistent ranking of kernel methods where \emph{alignf} yields the best results for all the three criteria. The second best method is the arc-cosine kernel, followed by the Gaussian kernel. All the three kernels (alignf, arc-cos and Gaussian) outperform the base setting used in \cite{Arora19}.
    }
    \label{fig:bestkernel_aggregated}
\end{figure*}


Figures \ref{fig:trainloss_alignf_cifar01}-\ref{fig:train_projections_loss_alignf_cifar01}, \ref{fig:trainloss_alignf_cifar23}-\ref{fig:train_projections_loss_alignf_cifar23}, and \ref{fig:trainloss_alignf_lsun}-\ref{fig:train_projections_loss_alignf_lsun} show the results of the \emph{alignf} embeddings with different dimensions on first two classes of CIFAR-10, second and third classes of CIFAR-10, and "kitchen" and "living room" classes of LSUN respectively. Similar to the experiments in \cite{CortesMohri2012}, we use a combination of Gaussian kernels (i.e., equation \eqref{eq:gaussian}) with changing the parameter $\gamma$. Specifically, we employ 7 different values for the parameter: $\gamma \in \{2^{-3}, 2^{-2}, \dots, 2^2, 2^3\}$. The experiments are repeated with 5 different random initializations. Figure \ref{fig:alignf_kernel_aggregated} illustrates the average of our results. The optimal kernel learned by the \emph{alignf} algorithm achieves the best results, as expected. Similar to Gaussian and arc-cosine kernels, the standard deviation of the results for alignf kernel was also significantly low. For instance, the standard deviation of training loss in different stages of training are in the order of $10^{-5}$ in the setting of figure \ref{fig:trainloss_alignf_cifar01}  and $10^{-4}$ in the settings of figures \ref{fig:trainloss_alignf_cifar23}  and  \ref{fig:trainloss_alignf_lsun} (when using alignf kernel embeddings with 10000 features). As in previous cases, we observe direct relations between a faster optimization, a better generalization and an improved overlap measure.  All the other embeddings such as the arc-cosine kernel designed as a proxy for neural embeddings could be viewed as trying to reach the performance of this \emph{optimal representation}.

\subsection{Discussion}
We have studied several  kernel embeddings to obtain more sophisticated features for the basic neural network model. 
In Figures \ref{fig:trainloss_bestkernel_cifar01}-\ref{fig:train_projections_bestkernel_cifar01}, \ref{fig:trainloss_bestkernel_cifar23}-\ref{fig:train_projections_bestkernel_cifar23} and \ref{fig:trainloss_bestkernel_lsun}-\ref{fig:train_projections_bestkernel_lsun} we compare the different kernels with the largest number of dimensions (i.e., when the number of features/components is $10,000$) which provides the most precise approximations.
We observe the consistency of fast optimization, low test error and high overlap between the labels vector and the top eigenvectors. In addition, we observe a consistent ranking of different kernel methods. The multiple kernel learning method in \cite{CortesMohri2012}  (called \emph{alignf}) yields the best results for all the three criteria and in all datasets. Indeed this is expected since it is explicitly optimized for overlap with data labels. 
All other kernels could be viewed as trying to reach the performance of this kernel. The second best method is the arc-cos kernel which mimics the computations in a neural network, and as expected, it performs better than a generic kernel like Gaussian (the kernel embeddings computed directly from a neural network are comparable to the arc-cos kernels, however they require a significantly larger runtime to be computed). Using the original data features yields the worst results, which indicates the importance of proper feature learning and inference for the task of interest.  

\begin{table}[th!]
    \centering
    \caption{Quantification of $  \sqrt{
  \frac{ 2\mathbf y^T(\mathbf H^\infty)^{-1}\mathbf y}{n}}$ (or $  \sqrt{
  \frac{ 2\mathbf y^T(\mathbf H(\mathcal{K})^\infty)^{-1}\mathbf y}{n}}$) for different kernels on first two classes of CIFAR-10 dataset ($n=10000$). The alignf method yields the smallest upper bound in the generalization error.}
    \begin{tabular}{l l r}
    
    \toprule
    Kernel & Dimension  & CIFAR-10 \\
          \midrule
          Gaussian & 10000 & 2.98 \\
          arc-cosine & 10000 & 2.59\\
          alignf & 10000 & 1.45 \\
          \multicolumn{2}{l}{No kernel} & 3.86\\
        \bottomrule
    \end{tabular}
    \label{tab:yHyCompareAll}
\end{table}

Finally, in Table \ref{tab:yHyCompareAll} we compare the theoretical upper bounds on the generalization errors of the different kernels when used with the first two classes of CIFAR-10. We observe a consistent ranking with the previous results. The \emph{alignf} method yields the best performance, and the arc-cos and the Gaussian kernels are the next choices, both better than the no-kernel case.

\section{Conclusions} We empirically explored the implications of  recent results of \cite{Arora19} for deeper networks, viewing previous layers as producing better representations of the input data. We studied different kernel and neural embeddings and showed that such representations benefit both optimization and generalization. We demonstrated the applicability of the overlap criteria in \cite{Arora19} for analysing the impacts of different kernels. 
In particular, we showed that a kernel optimally aligned to data yields the best results in terms of fast optimization, low generalization error, and large overlap between the top  eigenvectors and the labels vector, and that other embeddings achieved different approximations to this optimal representation.
By combining recent results connecting kernel embeddings to neural networks such as \cite{TKG18}, one may be able to extend the fine--grained theoretical results of \cite{Arora19} for two layer networks to deeper networks.

\bibliographystyle{spmpsci}
\bibliography{refs}

\end{document}